\documentclass[letterpaper]{article}

\usepackage[preprint]{aaai2027}

\usepackage[hyphens]{url}  
\usepackage{graphicx} 
\usepackage{natbib}  
\usepackage{caption} 
\usepackage{booktabs}
\usepackage{amsmath}
\usepackage{amssymb}
\usepackage{amsthm}
\usepackage{xcolor}
\usepackage{algorithm}
\usepackage{algorithmic}

\newcommand{\dS}{\mathit{dS}}
\newcommand{\dSbar}{\overline{\mathit{dS}}}

\newtheorem{proposition}{Proposition}

\title{CORAM: Coherent Orthogonal Rotation for Model Merging}

\author{
    Xinyi Sui\textsuperscript{\rm 1},
    Ziran Liu\textsuperscript{\rm 2},
    Nam Ling\textsuperscript{\rm 1},
    Wei Wang\textsuperscript{\rm 3},
    Wei Jiang\textsuperscript{\rm 3}
}

\affiliations{
    \textsuperscript{\rm 1}Santa Clara University\\
    \textsuperscript{\rm 2}Shanghai Institute for Mathematics and Interdisciplinary Sciences / Fudan University\\
    \textsuperscript{\rm 3}Futurewei Technologies, Inc.\\
    xsui@scu.edu, zliu@simis.cn, nling@scu.edu, rickweiwang@futurewei.com, wjiang@futurewei.com
}

\begin{document}

\maketitle

\begin{abstract}
Merging finetuned models combines specialized capabilities without joint training or access to the original data. 
Most methods operate by linear arithmetic in Euclidean weight space, which cannot carry the geometry of the update.
Orthogonal Model Merging (OrthoMerge) uses a single orthogonal transform for each weight matrix, but such a transform cannot change singular values. 
We propose CORAM, which partitions each target matrix into row slices, represents every expert slice by its singular value decomposition in the corresponding base-model SVD frame, and merges the task-specific factors on their corresponding manifolds. 
Because manifold averaging contracts the merged update, CORAM applies an amplification coefficient $\lambda=\kappa\hat{c}$. 
The scale $\hat{c}$ is estimated from the expert and merged update norms and is approximately $\sqrt{N}$ for $N$ experts with comparable update magnitudes. 
The restoration strength $\kappa$ is selected from the dispersion of expert updates without evaluating candidate merged models. 
This rule remains within $0.72$ points of the best swept value on all evaluated suites. 
CORAM also includes spread slicing to distribute highly updated rows across slices and a residual pathway for non-target layers. 
Across four suites covering three model families, 3B to 9B scales, and language and vision-language experts, CORAM improves over OrthoMerge by $0.25$ to $1.35$ points and matches or exceeds the strongest weight-space baselines.
\end{abstract}

\section{Introduction}

Model merging provides a training-free mechanism for consolidating independently finetuned capabilities, such as coding, mathematical reasoning, and multilingual understanding, into a single model without joint training or access to the original training data \citep{wortsman2022soups,ilharco2023task}. 
Most existing methods represent each finetuned model as a task vector in Euclidean weight space and combine these vectors through linear arithmetic \citep{ilharco2023task,yadav2023ties,yu2024dare}. 
Although simple and effective, this formulation ignores the geometric structure of weight updates. 
In particular, finetuning a weight matrix changes both its singular subspaces and its singular-value spectrum, which are not naturally represented by direct Euclidean addition.

Recent work addresses this limitation by performing model merging on structured geometric spaces. 
Orthogonal Model Merging (OrthoMerge) \citep{orthomerge}, for example, aligns finetuned weights using orthogonal transformations. 
However, it estimates a single transformation for an entire weight matrix. 
This matrix-level granularity can be restrictive because fine-tuning updates often concentrate in low-dimensional and heterogeneous subspaces \citep{aghajanyan2020intrinsicdimensionalityexplainseffectiveness,hu2021loralowrankadaptationlarge}. 
A single matrix-wide transformation may therefore mix strongly updated directions with largely unchanged ones.

\begin{figure}[t!]
    \centering
    \includegraphics[width=\columnwidth]{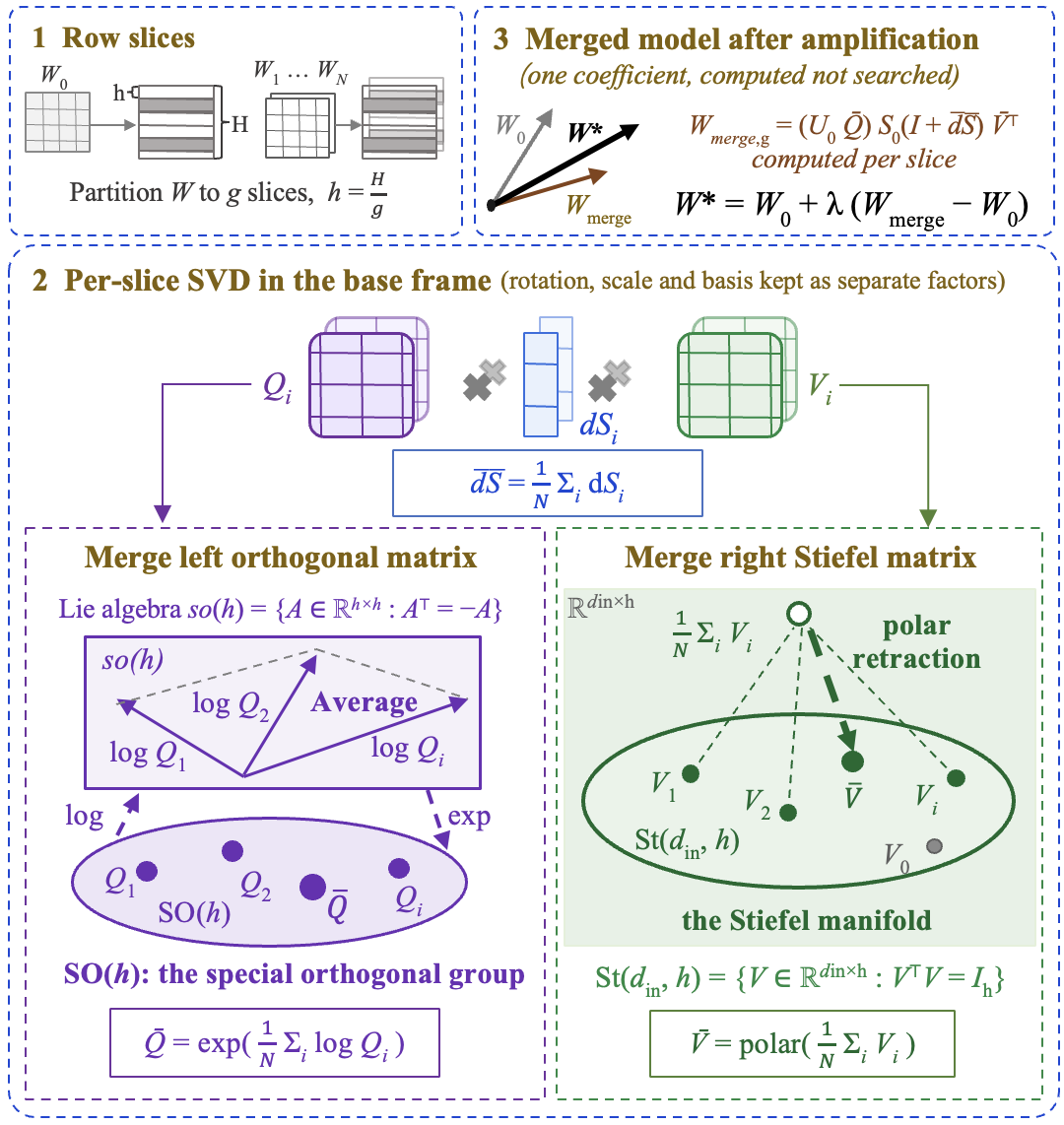}
   \caption{Overview of CORAM. Each expert slice is factored in the base frame into a rotation, a spectral shift $\dS_i = S_i S_0^{-1} - I$, and a right factor. Each factor is averaged in its own space. The coefficient $\lambda$ is given by Eq.~\eqref{eq:amplify}.}
    \label{fig:overview_coram}
\end{figure}
A finer representation is provided by CORA \citep{wang2026cora}, which parameterizes model adaptation using row-level weight slices. 
However, CORA addresses adaptation of an individual model rather than merging multiple independently finetuned models. Extending this representation to model merging raises two additional problems. First, the task-specific factors of each slice lie on different curved spaces and must be combined using geometry-compatible operations. Second, averaging on these spaces can substantially reduce the magnitude of the resulting update.

We introduce \textbf{CORAM}, a slice-level manifold method for model merging, illustrated in Fig.~\ref{fig:overview_coram}. 
CORAM partitions each target weight matrix into row slices and represents each expert slice through its SVD relative to the corresponding base-model slice. 
It merges three factors on their natural spaces: rotations are combined using a log-Euclidean mean on the special orthogonal group \citep{moakher2002rotmean,arsigny2006logeuclidean}, 
relative spectral shifts are combined through linear averaging, and right factors are aggregated using a polar mean on the Stiefel manifold \citep{edelman1998stiefel,6340355}. 
We additionally introduce a conflict-aware variant that masks, within each slice, neuron columns on which the experts exhibit inconsistent updates before aggregation. This adapts the conflict-handling mechanism of \citet{orthomerge} to the slice level.

A central challenge is that manifold averaging contracts the merged update toward the base model \citep{karcher2014riemanniancentermasscalled}. 
Without correction, this contraction can remove a substantial portion of the task-specific signal. 
CORAM compensates for the contraction using an amplification coefficient \(\lambda=\kappa\hat{c}\), where $\hat{c}$ estimates the contraction scale and $\kappa$ sets restoration strength. Importantly, neither term requires evaluating candidate merged models. 
The contraction scale $\hat{c}$ is computed from the norms of the expert and merged updates, and is approximated by $\sqrt{N}$ when the $N$ experts contribute updates of comparable magnitude, as established in Proposition~\ref{prop:sqrtn}.

The appropriate restoration strength $\kappa$ depends on how evenly the experts modify the base model. 
When expert updates have comparable magnitudes, near-full restoration is effective. 
When the magnitudes differ substantially, as observed for the strongest base model in our study, partial restoration performs better. 
CORAM distinguishes these two cases using a dispersion statistic computed from the expert updates without evaluating candidate merged models. 
The resulting rule selects the amplification coefficient without a sweep over $\lambda$ and remains within 0.72 score points of the best swept coefficient across all evaluated suites. 

The slice-level formulation motivates two complementary refinements. First, \emph{spread slicing} orders rows according to update magnitude and distributes them across slices, reducing the concentration of highly updated rows in a small number of slices. Second, a \emph{residual pathway} restores fine-tuning updates from layers outside the sliced merging targets, adapting the residual-decoupling principle of \citet{orthomerge} to the layer-level residuals. Their effectiveness also depends on the distribution of update magnitudes: spread slicing is beneficial by itself when expert updates are relatively uniform, whereas with strongly uneven updates it is most effective when combined with the residual pathway.

We evaluate CORAM on four heterogeneous model-merging suites spanning three model families, model sizes from 3B to 9B parameters, and both language and vision-language experts. Our main contributions are as follows:

\begin{itemize}

    \item \textbf{Search-free amplification.}
    We introduce the rule $\lambda=\kappa \hat{c}$, where the contraction scale $\hat{c}$ is estimated from expert and merged update norms and restoration strength $\kappa$ is selected using a zero-evaluation dispersion statistic. The predicted coefficient remains within 0.72 score points of the swept optimum on every evaluated suite. 
    
    \item \textbf{Slice-level manifold merging.} We merge per-slice SVD factors using geometry-compatible operations on the special orthogonal group, Euclidean spectral space, and the Stiefel manifold.

    \item \textbf{Geometry-motivated refinement.} We introduce spread slicing 
    and show how 
    its effectiveness changes with the dispersion of expert-update magnitudes.
    
    \item \textbf{Evaluation across heterogeneous settings.}
    Across four suites covering three model families, 3B--9B parameter scales, and language and vision-language tasks, CORAM improves over OrthoMerge by 0.25--1.35 score points under a common evaluation protocol and matches or exceeds the strongest evaluated weight-space baselines.    
\end{itemize}

\section{Related Work}

\paragraph{Euclidean weight-space merging.}
Most model-merging methods represent each finetuned expert as a task vector and combine the vectors in Euclidean weight space. Uniform averaging \citep{wortsman2022soups}, Fisher- or regression-weighted merging \citep{MatenaR22,jin2023dataless}, and task arithmetic \citep{ilharco2023task} differ mainly in how the updates are weighted.
TIES, DARE, and Localize-and-Stitch further reduce interference through sign agreement, sparsification, or parameter localization
\citep{yadav2023ties,yu2024dare,Localize2408}. 
These methods are simple and strong, but they treat a weight matrix as a flat vector and do not explicitly model its rotational and spectral structure.

\paragraph{Geometric and subspace merging.}
To move beyond direct weight arithmetic, existing methods align model units \citep{ainsworth2023git,ZipIt}, merge in tangent spaces \citep{TaskArithmetic2305-12827}, or construct subspaces from the SVD of task updates \citep{tsvm,stoica2025model,marczak2025no}. Closest to
CORAM, OrthoMerge averages one orthogonal transform per weight matrix and merges the remaining residual conventionally \citep{orthomerge}. An orthogonal transform cannot change singular values, so the spectral part of the update remains in the residual. CORA provides a finer per-slice SVD representation, but studies model adaptation rather than
multi-expert merging \citep{wang2026cora}. CORAM uses this per-slice representation for model merging and averages the rotation, relative spectral shift, and right factor in their spaces. Its conflict-aware variants and residual pathway adapt the conflict handling and residual decoupling of OrthoMerge to slices and non-target layers.

\paragraph{Coefficient selection in model merging.}
Merging on these structured spaces raises a further question: how to set the scale of the merged update. Task arithmetic uses a fixed constant \citep{ilharco2023task}, and benchmark protocols select one value per suite \citep{mergebench}. AdaMerging instead optimizes per-layer coefficients at test time \citep{YangW00G0T24}, while Model Stock derives a geometric rule for models finetuned on the same task \citep{jang2025modelstockneedjust}. CORAM relates the coefficient to the contraction of the geometric merge. The contraction scale is measured from the expert and merged update norms, and the restoration strength is selected from the dispersion of expert-update magnitudes. This choice is motivated by prior observations that merging behavior changes with the base model \citep{mergebench,yadav2025what}, and avoids a per-benchmark sweep over $\lambda$.

\section{Method}
\label{sec:method}

\subsection{Slice Representation in the Base Frame}
\label{sec:slicerep}

Following CORA \citep{wang2026cora}, we partition each target weight matrix $W \in \mathbb{R}^{d_\mathrm{out}\times d_\mathrm{in}}$ into $G = d_\mathrm{out}/h$ row slices of height $h$, $W^{(g)} = W[gh{:}(g{+}1)h,\,:]$, and we work in the base frame. Each base slice is factored once as $W_0^{(g)} = U_0\,S_0\,V_0^{\top}$. For each expert $i \in \{1,\dots,N\}$ and each slice, we express the expert weights relative to this frame using three quantities: a rotation $Q_i \in SO(h)$ aligning the expert left singular basis with $U_0$, obtained by the polar factor of $U_0^{\top}U_i$; a relative spectral shift $\dS_i = S_i S_0^{-1} - I$; and the right factor $V_i$. Before forming $Q_i$, we resolve the sign ambiguity of each singular direction jointly against the base pair $(U_0,V_0)$: each $(u,v)$ pair is flipped together, leaving $W_i$ unchanged. The triple $(Q_i, \dS_i, V_i)$ lies on a product of structured spaces: the special orthogonal group $SO(h)$, the relative spectral shifts, and the Stiefel manifold $\mathrm{St}(d_\mathrm{in}, h)$ \citep{b3659c5c-07ee-3086-99fb-92b98601b32d}, and this structure dictates how the factors should be averaged.

\subsection{Per-Slice Manifold Merging}
\label{sec:manifoldmerge}

Given the per-expert slice triples, CORAM merges each factor by the mean native to its space (Fig.~\ref{fig:overview_coram}, panel 2):

\noindent\emph{\textbf{Rotations.}} We average $\{Q_i\}$ by the log-Euclidean mean on $SO(h)$ \citep{moakher2002rotmean,arsigny2006logeuclidean}, mapping each rotation to the Lie algebra, averaging, and mapping back,
\begin{equation}
\bar{Q} \;=\; \exp\!\Big(\tfrac{1}{N}\textstyle\sum_{i}\log Q_i\Big).
\label{eq:lem}
\end{equation}
Because $\mathfrak{so}(h)$ is a linear space, the operation inside the exponential is an ordinary mean, while the nonlinearity enters only through the logarithmic and exponential maps.

\noindent\emph{\textbf{Spectral shifts.}} The relative spectral shifts are averaged linearly, $\dSbar = \sum_i \tfrac{1}{N} \dS_i$.

\noindent\emph{\textbf{Right factors.}} We average $\{V_i\}$ by the polar mean on the Stiefel manifold \citep{edelman1998stiefel,6340355}, forming the weighted Euclidean sum and projecting back onto the manifold via the polar decomposition $\bar{V} = \mathrm{polar}\!\big(\textstyle\sum_i \tfrac{1}{N} V_i\big)$.

The merged slice is reconstructed in the base frame as
\begin{equation}
W_{\mathrm{merge},g} \;=\; \big(U_0\,\bar{Q}\big)\,
S_0\,\big(I+\dSbar\big)\,\bar{V}^{\top}
\;+\; \bar{\eta}^{(g)},
\label{eq:reconstruct}
\end{equation}
where $\bar{\eta}^{(g)}$ is a conflict residual, defined in the conflict-aware variants below, and is zero when no masking is applied.
Stacking the merged slices yields $W_\mathrm{merge}$ for the matrix. Applying the same procedure to all target matrices produces the geometric merge of the model.

\subsection{Conflict-Aware Variants}
\label{sec:conflict}
Experts can disagree sharply on individual neurons, and averaging across such disagreements dilutes every expert's contribution. Adapting the conflict-handling idea of Orthogonal Model Merging \citep{orthomerge} to the slice level, our conflict-aware variants detect, per slice and per column, the experts whose update direction opposes the consensus: with $\tau_i = W_i^{(g)} - W_0^{(g)}$ and $\bar{\tau} = \tfrac{1}{N}\sum_i \tau_i$, column $j$ of expert $i$ is flagged when $\cos\!\big(\tau_i[:,j],\, \bar{\tau}[:,j]\big)$ is negative. The flagged columns split each expert update in two parts: the part that enters the manifold means above, and the complement, which is either discarded or averaged across experts and added back to Eq.~\eqref{eq:reconstruct} as $\bar{\eta}^{(g)}$. Combining these choices yields the three variants used in our experiments:
\begin{itemize}
    \item The non-flagged columns are merged on the manifolds, and the flagged ones are discarded.
    \item The non-flagged columns are merged on the manifolds, and the flagged ones return as $\bar{\eta}^{(g)}$.
    \item The flagged columns are merged on the manifolds, and the non-flagged ones return as $\bar{\eta}^{(g)}$.
\end{itemize}
\noindent The fourth combination, merging only the flagged columns and discarding the rest, would throw away the signal on which the experts agree, and we do not use it. 
For the first two variants, flagged columns are masked out of that expert’s contribution before the slice is factored, so the per-slice manifold means above are taken over non-conflicting evidence only. We refer to the plain method as CORAM and collectively denote the conflict-aware variants by CORAM-C. Both method families are evaluated throughout.

\subsection{Amplitude Restoration}
\label{sec:amplitude}

Merging on curved spaces introduces an effect absent from Euclidean merging. The manifold mean \emph{contracts} the merged update toward the base, i.e., $\|W_\mathrm{merge} - W_0\|$ is substantially smaller than the typical expert update $\|W_i - W_0\|$. The merged direction is preserved, but its magnitude is reduced. CORAM therefore applies a single global amplification,
\begin{equation}
W^{*} \;=\; W_0 + \lambda\,\big(W_\mathrm{merge} - W_0\big),
\qquad \lambda = \kappa\,\hat{c},
\label{eq:amplify}
\end{equation}
where the scale $\hat{c}$ and strength $\kappa$ are determined without evaluating candidate merged models.

\noindent\emph{\textbf{Contraction scale.}} The contraction is directly measurable from quantities already computed during merging:
\begin{equation}
\hat{c} \;=\; c_{\mathrm{RMS}} \;=\;
\frac{\mathrm{RMS}_i\,\|\Delta W_i\|}{\|\Delta W_\mathrm{merge}\|},
\qquad \Delta W \equiv W - W_0,
\label{eq:crms}
\end{equation}
the ratio between the typical expert update magnitude and the merged one. No task evaluation is involved. In the ideal case, this rate has a closed form:

\begin{proposition}[Cancellation scale]
\label{prop:sqrtn}
Let $\tau_1,\dots,\tau_N$ be slice updates with $\mathbb{E}\langle\tau_i,\tau_j\rangle = 0$ for $i\neq j$ and $\|\tau_i\| = \|\tau\|$ for all $i$. Then $\mathbb{E}\,\big\|\tfrac{1}{N}\sum_i \tau_i\big\|^2 = \|\tau\|^2/N$, so restoring the mean to the typical expert magnitude requires amplification by $\sqrt{N}$.
\end{proposition}

The proposition follows from the random-vector cancellation law. It applies directly to the rotation average in the linear space $\mathfrak{so}(h)$ and to the Euclidean average of the relative spectral shifts (supplementary material). The additional effect of the Stiefel mean is included in
$c_{\mathrm{RMS}}$, which extends the $\sqrt{N}$ scale to correlated or uneven updates. Across four low-dispersion settings, $c_{\mathrm{RMS}}=2.09/2.03/1.72/2.09$, compared with $\sqrt{N}=2.24/2.24/1.73/2.24$, with a difference below $10\%$ in each case. We therefore use $\lambda=\kappa\sqrt{N}$ as the default prescription and retain $c_{\mathrm{RMS}}$ as the direct measurement of contraction.

\noindent\emph{\textbf{Restoration strength.}} 
The remaining constant $\kappa$ is determined empirically. 
Sweeps on every base place its optimum in one of two cases: a low-dispersion case, with comparable expert update magnitudes and near-full restoration ($\kappa\approx1.15$), and a high-dispersion case, with strongly uneven update magnitudes and partial restoration ($\kappa\approx0.5$). 
We distinguish these cases \emph{before task evaluation} using the dispersion statistic,
\begin{equation}
D =\frac{\max_i r_i}{\min_i r_i},
\quad r_i = \frac{\|\Delta W_i\|}{\|W_0\|},
\label{eq:dispersion}
\end{equation}
computed from the same cached quantities as Eq.~\eqref{eq:crms}. 
Small $D$ selects the low-dispersion case, whereas large $D$ selects the high-dispersion case. The amplification coefficient is therefore set from Proposition~\ref{prop:sqrtn} and $D$, without a per-benchmark sweep over $\lambda$. We also tested alternative closed-form choices over $\lambda\in[1,4]$, as reported in the supplementary material.

\section{Geometry-Derived Refinements}
\label{sec:refinements}

The slice-level formulation supports two additional refinements. Both are optional and are evaluated separately and together. Their effects depend on the distribution of expert-update magnitudes. 
Spread slicing is effective by itself when the update magnitudes are comparable. When the update magnitudes differ substantially, it works better together with the residual pathway.

\subsection{Spread Slicing}
\label{sec:spread}


Contiguous slicing follows the row order of the pretrained matrix. If heavily updated rows cluster in a small number of slices, these slices may have poorly conditioned SVDs while the remaining slices contain little update signal. Spread slicing distributes the important rows more evenly. 
Before slicing, we score every row $r$ of every expert by its relative update magnitude, $s_i(r) = \|\Delta W_i[r,:]\| / \|W_0[r,:]\|$, and we summarize the scores of a row across the experts by their largest value $s_{(1)}$, the second largest $s_{(2)}$, their mean, and their variance. A priority is formed from these quantities, and the rows are then handed out to slices in descending priority. Four settings are available:
\begin{itemize}
\item \textbf{Mean.} The priority is the mean of the scores, which weights all experts equally.
\item \textbf{Energy.} The priority is $s_{(1)}$, so the rows that some expert updates most strongly are spread first.
\item \textbf{Variance.} The priority is
$\operatorname{Var}_i[s_i(r)]\,s_{(1)}$.
The variance is large when experts disagree on how much to update the row, while $s_{(1)}$ discounts rows with little update signal.
\item \textbf{Owner balance.} The priority is $(s_{(1)}-s_{(2)})\,s_{(1)}$, which is large when an expert dominates the row and that row also changes a lot. 
\end{itemize}
The first three settings assign rows to slices in round-robin order. Owner balance assigns each row to the slice with the smallest accumulated priority, and breaks ties to prevent rows dominated by the same expert from concentrating in one slice. After merging, the permutation is inverted to restore the original row order. This permutation changes only how rows are grouped into slices and does not change the function computed by any expert. The same permutation is applied to all experts, so the merge remains well defined. We denote this configuration by CORAM$+$SS.

\subsection{Layer-Level Residual Pathway}
\label{sec:residual}

The slice geometry covers the target linear layers, including the attention and MLP projections. Embeddings and normalization parameters are not included, so a purely geometric merge omits their fine-tuning updates. Following the orthogonal and residual decoupling idea of \citet{orthomerge}, we merge the non-target residuals $\rho_i = W_i - W_0$ using a conventional weight-space method and add the result as a separate patch.

We use task arithmetic \citep{ilharco2023task}, TIES \citep{yadav2023ties}, or Task Singular Vectors \citep{tsvm} for the residual merge. For Task Singular Vectors, we follow the distributed-TSVM implementation of \citet{orthomerge}. We denote this configuration by CORAM$+$RP. The final model combines the amplified geometric merge with the unscaled residual patch,

\begin{equation}
W^{*} \;=\; W_0 +
\begin{cases}
\lambda\,(W_\mathrm{merge}-W_0) & \text{target layers},\\
\bar{\rho} & \text{non-target},
\end{cases}
\label{eq:assembly}
\end{equation}
where $\lambda$ follows the amplitude-restoration rule: the amplification corrects the contraction of the \emph{manifold} mean only, while the residual merger is Euclidean and needs no correction. When both refinements are enabled, CORAM$+$SS$+$RP applies spread slicing before geometric merging and adds the residual patch afterward.

\section{Experiments}
\label{sec:experiments}
\vspace{-.2em}
\begin{table*}[t!]
\centering
\small
\renewcommand{\arraystretch}{0.90}
\setlength{\tabcolsep}{4pt}
\begin{tabular}{lccccc|cc|cc|c}
\toprule
& \multicolumn{5}{c|}{In-domain} & \multicolumn{2}{c|}{In-domain Avg}
& \multicolumn{2}{c|}{Out-of-domain} & \\
Method & MATH500 & HEval$+$ & SciQA & CSQA & SIQA & $\kappa$-rule & best $\lambda$
& M-ARC & AGIEval & OOD Avg \\
\midrule
Llama-3.1-8B & 17.80 & 21.52 & 71.40 & 70.60 & 48.11 & \multicolumn{2}{c|}{45.89} & 33.02 & 30.27 & 31.65 \\
Task-specific FT & 19.00 & 38.54 & 91.82 & 82.47 & 56.81 & \multicolumn{2}{c|}{57.73} & -- & -- & -- \\
\midrule
Linear & 23.40 & 32.26 & 83.59 & 76.82 & 51.84 & \multicolumn{2}{c|}{53.58} & 34.41 & 32.27 & 33.34 \\
Task Arithmetic & \textbf{24.80} & 37.20 & 86.29 & 79.52 & 53.33 & \multicolumn{2}{c|}{56.23} & 34.97 & 32.68 & 33.83 \\
TIES & 21.40 & \textbf{40.43} & 82.87 & 78.30 & 54.04 & \multicolumn{2}{c|}{55.41} & 34.40 & 32.53 & 33.47 \\
DARE-TIES & 19.20 & 40.30 & 82.24 & 78.54 & 54.45 & \multicolumn{2}{c|}{54.95} & 33.50 & 32.24 & 32.87 \\
OrthoMerge (OFT) & 24.20 & 37.87 & 87.72 & 80.75 & 55.12 & \multicolumn{2}{c|}{57.13} & 35.15 & \textbf{33.13} & \textbf{34.14} \\
\midrule
CORAM & 21.80 & \textbf{40.43} & 88.31 & \textbf{81.24} & \textbf{56.81} & \textbf{57.72} & 57.98\,{\scriptsize($h{=}16$, $\lambda{=}2.40$)} & \textbf{35.21} & 32.53 & 33.87 \\
CORAM-C & 20.60 & 38.41 & \textbf{88.44} & \textbf{81.24} & \textbf{56.81} & 57.10 & 57.74\,{\scriptsize($\lambda{=}2.25$)} & 35.15 & 32.61 & 33.88 \\
CORAM-C$+$SS & 21.40 & 39.88 & 88.40 & 81.08 & 56.70 & 57.49 & \textbf{58.21}\,{\scriptsize($\lambda{=}2.30$)} & 35.12 & 32.17 & 33.65 \\
\bottomrule
\end{tabular}
\par\vspace{7pt}
\begin{tabular}{lccccc|cc|cc|c}
\toprule
& \multicolumn{5}{c|}{In-domain} & \multicolumn{2}{c|}{In-domain Avg}
& \multicolumn{2}{c|}{Out-of-domain} & \\
Method & Instr. & Math & Coding & Multi. & Safety
& $\kappa$-rule & best $\lambda$ & MMLU$^{\dagger}$ & AGIEval & OOD Avg \\
\midrule
Llama-3.2-3B & 7.58 & 28.51 & 27.44 & 40.72 & 31.41 & \multicolumn{2}{c|}{27.18} & 56.79 & 24.01 & 40.40 \\
Task-specific FT & 39.56 & 69.83 & 44.33 & 41.73 & 80.46 & \multicolumn{2}{c|}{55.18} & -- & -- & -- \\
\midrule
Linear & 9.80 & 40.86 & 37.15 & 42.22 & 40.21 & \multicolumn{2}{c|}{34.05} & \textbf{57.54} & 25.88 & 41.71 \\
Task Arithmetic & 18.30 & 45.49 & 40.18 & \textbf{42.47} & 44.90 & \multicolumn{2}{c|}{38.27} & 57.50 & 26.82 & \textbf{42.16} \\
TIES & 22.18 & 50.04 & 39.40 & 42.03 & 41.93 & \multicolumn{2}{c|}{39.12} & 56.76 & \textbf{27.08} & 41.92 \\
DARE-TIES & 31.61 & \textbf{56.94} & 39.01 & 41.22 & 48.50 & \multicolumn{2}{c|}{43.45} & 55.94 & 25.55 & 40.74 \\
OrthoMerge (TSV-M+C) & 20.15 & 55.57 & 41.76 & 42.28 & \textbf{50.90} & \multicolumn{2}{c|}{42.13} & 57.42 & 26.87 & 42.14 \\
\midrule
CORAM$+$SS$+$RP & 31.24 & 52.99 & 42.12 & 41.72 & 48.40 & 43.30 & 43.30\,{\scriptsize($\lambda{=}2.60$)} & 56.02 & 25.60 & 40.81 \\
CORAM-C$+$RP & \textbf{31.79} & 53.83 & 40.79 & 41.56 & 48.65 & 43.32 & 43.32\,{\scriptsize($\lambda{=}2.60$)} & 56.18 & 25.60 & 40.89 \\
CORAM-C$+$SS$+$RP & 30.50 & 52.84 & \textbf{42.40} & 41.70 & 49.94 & \textbf{43.48} & \textbf{43.48}\,{\scriptsize($\lambda{=}2.60$)} & 56.17 & 25.67 & 40.92 \\
\bottomrule
\end{tabular}
\caption{\textbf{Top--T1}: merging five orthogonally finetuned Llama-3.1-8B experts. \textbf{Bottom--T2}: merging five fully finetuned Llama-3.2-3B experts from MergeBench. Higher is better. Per column, the best result among merging methods is in bold (base-model and task-specific-FT reference rows excluded). CORAM-C denotes the conflict-aware variants (masking neurons where experts disagree). $+$SS adds spread slicing (a task-informed row permutation) and $+$RP the residual pathway (conventional merge of the non-target layers). CORAM rows evaluate the search-free $\kappa$-rule checkpoint (grid $\lambda{=}2.60$ for both suites). ``best $\lambda$'' is the sweep optimum, for reference (see Setup).}
\label{tab:t1t2}\vspace{-1.5em}
\end{table*}

\subsection{Setup}
\label{sec:setup}

\paragraph{Suites.} 
We evaluate CORAM on four suites covering three model families, model sizes from 3B to 9B, and both language and vision-language experts. 

\textbf{T1} contains five orthogonally finetuned Llama-3.1-8B experts \citep{Controlling230607280} released by \citet{orthomerge}. We evaluate them on MATH500 \citep{Measuring210303874}, HumanEval$+$ \citep{IsYourCode230501210}, ScienceQA \citep{LearntoExplain220909513}, CommonsenseQA \citep{CommonsenseQA181100937}, and Social-IQA \citep{SapRCBC19}.

\textbf{T2} and \textbf{T4} each contain five fully finetuned experts for instruction following, mathematics, coding, multilingual tasks, and safety. The experts are based on Llama-3.2-3B and Gemma-2-9B and are obtained from MergeBench \citep{mergebench}. We follow the task suites, evaluation protocol, and hyperparameter settings of MergeBench.

\textbf{T3} contains three Qwen2.5-VL-7B-Instruct experts for spatial reasoning, OCR, and medical multimodal question answering. We follow the vision-language setup of \citet{orthomerge} and evaluate on MMSI-Bench \citep{MMSI-Bench}, EmbSpatial \citep{DuWLHW24}, MMMU-Med \citep{YueNZ0LZSJRSWYY24}, PathVQA \citep{he2020pathvqa30000questionsmedical}, OCRBench \citep{Liu_2024}, and CharXiv \citep{CharXiv}. We use the multiple-choice subset of MMMU-Med. CharXiv is scored using its official GPT judge, gpt-4o-2024-05-13. Our evaluation pipeline obtains 69.50 for the best variant of \citet{orthomerge}, compared with the reported score of 69.90.

\paragraph{Cost.} 
CORAM uses slice SVDs that are computed once and cached for subsequent merges. 
The geometric merge requires 0.5 to 0.8 GPU-hours for each configuration.
The evaluation cost of a sweep over $\lambda$ is 
one to two orders of magnitude larger than the merge cost. 
Detailed storage and runtime statistics are provided in the supplementary material.

\paragraph{Out-of-domain evaluation.} 
We evaluate the retention of general capabilities using the out-of-domain (OOD) tasks adopted by \citet{orthomerge}. 
For T1, we use M-ARC and AGIEval \citep{AGIEval230406364}. For T2 and T4, we use MMLU$^{\dagger}$ \citep{HendrycksBBZMSS21} and AGIEval, with the mathematics and coding subsets removed from MMLU. 
For T3, we use IFEval \citep{zhou2023instructionfollowingevaluationlargelanguage} and MMBench \citep{MMBench}. 
All OOD tasks are evaluated zero-shot using the same evaluation harness. The specific differences are provided in the supplementary material.

\begin{table*}[t!]
\centering
\small
\renewcommand{\arraystretch}{0.90}
\setlength{\tabcolsep}{3.5pt}
\begin{tabular}{lcccccc|cc|cc|c}
\toprule
& \multicolumn{6}{c|}{In-domain} & \multicolumn{2}{c|}{In-domain Avg}
& \multicolumn{2}{c|}{Out-of-domain} & \\
Method & MMSI & EmbSp. & MMMU$_\mathrm{Med}$ & PathVQA & OCRB. & CharXiv
& $\kappa$-rule & best $\lambda$ & IFEval & MMB. & OOD Avg \\
\midrule
Qwen2.5-VL-7B-It. & 27.80 & 69.97 & 53.10 & 66.30 & 84.70 & 67.20 & \multicolumn{2}{c|}{61.51} & 63.03 & 83.93 & 73.48 \\
Task-specific FT & 32.60 & 70.58 & 55.17 & 66.81 & 85.00 & 72.50 & \multicolumn{2}{c|}{63.78} & -- & -- & -- \\
\midrule
Linear & 29.20 & 71.29 & 55.17 & \textbf{68.47} & 84.80 & 67.30 & \multicolumn{2}{c|}{62.71} & 58.23 & \textbf{84.19} & 71.21 \\
Task Arithmetic & 29.10 & 71.07 & 55.86 & 68.38 & 84.60 & 66.10 & \multicolumn{2}{c|}{62.52} & \textbf{59.33} & \textbf{84.19} & \textbf{71.76} \\
TIES & 32.10 & 71.54 & 57.93 & 68.44 & 82.80 & 69.40 & \multicolumn{2}{c|}{63.70} & 54.53 & 84.02 & 69.27 \\
DARE-TIES & 32.10 & 71.76 & 58.62 & 66.98 & 80.80 & \textbf{69.60} & \multicolumn{2}{c|}{63.31} & 51.02 & 82.82 & 66.92 \\
OrthoMerge (TIES+C) & 32.30 & 71.76 & 56.55 & 68.14 & 83.10 & 69.50 & \multicolumn{2}{c|}{63.56} & 54.53 & 83.68 & 69.10 \\
\midrule
CORAM & 33.20 & 71.59 & 60.00 & 66.69 & \textbf{85.40} & 67.00 & 63.98 & 64.10\,{\scriptsize($\lambda{=}1.60$)} & 53.42 & 83.25 & 68.33 \\
CORAM$+$RP & 33.20 & \textbf{72.17} & \textbf{60.69} & 67.22 & 84.90 & 68.10 & \textbf{64.38} & \textbf{64.38}\,{\scriptsize($\lambda{=}2.00$)} & 54.53 & 82.99 & 68.76 \\
CORAM-C$+$RP & \textbf{33.70} & 72.03 & 59.31 & 67.37 & 84.90 & 67.90 & 64.20 & 64.26\,{\scriptsize($\lambda{=}1.75$)} & 52.68 & 83.16 & 67.92 \\
\bottomrule
\end{tabular}
\caption{\textbf{T3}: merging three Qwen2.5-VL-7B-Instruct vision--language experts. Layout and bolding as in Table~\ref{tab:t1t2}. $\kappa$-rule grid point $\lambda{=}2.00$. CORAM-C denotes the conflict-aware variants (masking neurons where experts disagree). $+$RP adds the residual pathway (conventional merge of the non-target layers). The OrthoMerge out-of-domain entries are obtained from our zero-shot re-evaluation.}
\label{tab:t3}
\end{table*}
\begin{table*}[t!]
\centering
\small
\renewcommand{\arraystretch}{0.90}
\setlength{\tabcolsep}{4pt}
\begin{tabular}{lccccc|cc|cc|c}
\toprule
& \multicolumn{5}{c|}{In-domain} & \multicolumn{2}{c|}{In-domain Avg}
& \multicolumn{2}{c|}{Out-of-domain} & \\
Method & Instr. & Math & Coding & Multi. & Safety
& $\kappa$-rule & best $\lambda$ & MMLU$^{\dagger}$ & AGIEval & OOD Avg \\
\midrule
Gemma-2-9B & 14.23 & 69.83 & 43.37 & 54.63 & 34.39 & \multicolumn{2}{c|}{43.29} & 70.41 & 38.01 & 54.21 \\
Task-specific FT & 65.06 & 79.76 & 58.51 & 55.91 & 76.00 & \multicolumn{2}{c|}{67.05} & -- & -- & -- \\
\midrule
Linear & 27.17 & 81.05 & 51.52 & \textbf{53.79} & 59.31 & \multicolumn{2}{c|}{54.57} & \textbf{67.88} & 37.75 & 52.81 \\
Task Arithmetic & \textbf{27.54} & 82.34 & 51.65 & 51.31 & 54.26 & \multicolumn{2}{c|}{53.42} & 64.44 & 35.70 & 50.07 \\
TIES & 24.03 & \textbf{82.94} & 46.60 & 45.64 & 52.84 & \multicolumn{2}{c|}{50.41} & 60.60 & 34.92 & 47.76 \\
DARE-TIES & 18.11 & 76.57 & 20.22 & 34.23 & 49.12 & \multicolumn{2}{c|}{39.65} & 42.18 & 27.10 & 34.64 \\
OrthoMerge (TA+C) & 26.43 & 81.27 & 52.83 & 53.78 & 59.37 & \multicolumn{2}{c|}{\textbf{54.74}} & 67.80 & 37.85 & \textbf{52.83} \\
\midrule
CORAM-C & 27.36 & 81.27 & 51.62 & 53.52 & 57.83 & 54.32 & 54.32\,{\scriptsize($\lambda{=}1.10$)} & 67.22 & \textbf{38.03} & 52.63 \\
CORAM$+$RP & 26.06 & 81.58 & 51.84 & 53.59 & \textbf{59.73} & 54.56 & 54.84\,{\scriptsize($\lambda{=}1.25$)} & 67.30 & 37.28 & 52.29 \\
CORAM$+$SS$+$RP & 26.25 & 81.80 & \textbf{53.10} & 53.66 & 58.28 & 54.62 & \textbf{54.99}\,{\scriptsize($\lambda{=}1.25$)} & 67.59 & 37.62 & 52.60 \\
\bottomrule
\end{tabular}
\caption{
\textbf{T4}: merging five fully finetuned Gemma-2-9B experts from MergeBench with uneven update magnitudes ($D{\approx}16$). The rule selects $\kappa{=}0.5$, corresponding to $\lambda{=}1.10$. Layout and bolding follow Table~\ref{tab:t1t2}. CORAM-C masks neurons where experts disagree. $+$SS adds a task-informed row permutation, and $+$RP merges the non-target layers. 
}
\label{tab:t4}\vspace{-.5em}
\end{table*}

\paragraph{Baselines and CORAM configurations.} 
We implement linear averaging \citep{wortsman2022soups}, task arithmetic \citep{ilharco2023task}, TIES \citep{yadav2023ties}, and DARE-TIES \citep{yu2024dare,yadav2023ties} as per-tensor algorithms following their original definitions. We use the MergeBench coefficient settings and the baseline set of \citet{orthomerge}. For Orthogonal Model Merging \citep{orthomerge}, we evaluate all released variants under our protocol and report the strongest result for each suite. On T2, our reproduction obtains 42.13, compared with the published 42.07. 
All table comparisons use the same protocol. Reproduction details are provided in the supplementary material.

The base-model and task-specific finetuning results for T1, T2, and T3 are taken from \citet{orthomerge}. The corresponding results for T4 are measured using our evaluation harness. CORAM denotes the plain method, and CORAM-C denotes the conflict-aware variants. CORAM$+$SS includes spread slicing, while CORAM$+$RP includes the residual pathway. The main tables report three representative configurations for each suite. The complete component combinations are reported in the supplementary material.

We use a slice height of $h\!=\!8$ and $h\!=\!16$ in experiments. 
Each CORAM row evaluates one checkpoint selected without a sweep over $\lambda$. The task scores, OOD scores, and $\kappa$-rule average are obtained from the model merged using $\lambda\!=\!\kappa\sqrt{N}$. $\kappa$ is selected using the dispersion statistic $D$ in Eq.~\eqref{eq:dispersion}. The best $\lambda$ column is the only result obtained from a sweep. It reports the in-domain average at the best value for each suite and measures the difference between the selected coefficient and the sweep optimum. The detailed configuration corresponding to each CORAM row is reported in the supplementary material.

\subsection{Main Results}
\label{sec:mainresults}\vspace{-.5em}

Tables~\ref{tab:t1t2}--\ref{tab:t4} report the main comparison. \emph{First}, CORAM outperforms every OrthoMerge variant on all four suites, by $+1.08$ (T1), $+1.35$ (T2), $+0.82$ (T3), and $+0.25$ (T4) at the respective best configurations, all measured under our single harness against our reproduction of OrthoMerge (see Setup). \emph{Second}, against the strongest weight-space baselines CORAM leads on T1, T3, and T4, and matches DARE-TIES on T2 (43.48 vs.\ 43.45). No baseline is consistently competitive: the method strongest on any one suite trails CORAM by 1.1--15.3 points on the others. Retention tells the same story: on T4 the aggressive DARE-TIES baseline collapses out-of-domain (34.64 vs.\ our 52.60), while CORAM's OOD averages stay within the band of the mildest mergers on every suite. \emph{Third}, the search-free $\kappa$-rule recovers near-peak accuracy throughout: across all suites and configurations, its gap to the swept-$\lambda$ optimum is at most 0.72 points, including on T4, where the rule selects the \emph{high-dispersion} case ($\kappa{\approx}0.5$) purely from the zero-evaluation dispersion $D$. It removes the need for a per-suite $\lambda$ sweep of ten or more grid points, each requiring a full-suite evaluation.

\subsection{The Amplification Dichotomy}
\label{sec:analysis}
\paragraph{Two cases.} 
T4 evaluates the rule on a different model family and the largest model considered. The value of $D$ is computed before task evaluation. Across the five bases (see supplementary material), $D$ separates the models into low- and high-dispersion cases. Qwen2.5-VL-7B, Gemma-2-2B, Llama-3.2-3B, and Llama-3.1-8B have $D\in[1.3,3.5]$, while Gemma-2-9B has $D\approx16$. The optimal $\kappa$, measured by $\lambda$ sweeps on every base, follows the same division. It lies between approximately $0.9$ and $1.2$ in the low-dispersion case and near $0.5$ in the high-dispersion case.

We therefore adopt two shared constants, $\kappa{=}1.15$ and $\kappa{=}0.5$, fixed once across all suites rather than fitted per suite. 
Because the sweep optima are flat, this choice differs from the per-suite optimum by at most $0.72$ points. Using $\kappa=1.10$ does not change any comparison in the tables. The supplementary material reports the per-configuration ranges of $\kappa_{\mathrm{opt}}$, the relation between $\kappa$ and $D$, and a multi-seed evaluation of the main-table checkpoints. The value used in the high-dispersion case is calibrated on the only available base with a large value of $D$, as discussed in the limitations. Our results support these two cases but do not establish how $\kappa$ behaves for intermediate values of $D$.

\paragraph{Relation to base model strength.} The two cases are not explained by model family or fine-tuning method. Within Gemma-2, the 2B base has $D=1.67$, while the 9B base has $D=16$, so the selected $\kappa$ changes with scale within the same architecture. The 9B math expert is trained with GRPO, while the others use SFT. Excluding the math expert gives $D=16.0$ for Gemma-2-9B and $D=1.67$ for Gemma-2-2B under matched fine-tuning methods. The difference is therefore more closely associated with base-model strength, consistent with observations that merging behaves differently on stronger bases \citep{mergebench,yadav2025what}. The statistic $D$ measures this difference before merging.

\paragraph{Explanation.} 
Proposition~\ref{prop:sqrtn} assumes nearly orthogonal updates with comparable norms. A small value of $D$ indicates that the update norms are comparable, so near-full restoration is appropriate. At $D\approx16$, the update magnitudes differ substantially and the equal-norm assumption no longer holds. Full restoration then amplifies the merged update too strongly, and a value near $\kappa=0.5$ performs better.
The T4 results in Table~\ref{tab:t4} show that $D$ distinguishes the two cases before task evaluation.

\begin{table}[t!]
\centering
\small
\renewcommand{\arraystretch}{0.90}
\setlength{\tabcolsep}{5pt}
\begin{tabular}{lcccc}
\toprule
Configuration & T1 & T2 & T3 & T4 \\
\midrule
CORAM            & 57.72 & 42.75 & 63.95 & 54.49 \\
\;$+$RP          & n/a   & 42.31 & \textbf{64.38} & 54.56 \\
\;$+$SS          & 57.84    & 42.93 & 63.96 & 54.15 \\
\;$+$SS$+$RP     & n/a   & 43.30 & 64.27 & \textbf{54.71} \\
\midrule
CORAM-C          & 57.60 & 42.96 & 64.02 & 54.32 \\
\;$+$RP          & n/a & 43.32 & 64.20 & 54.37 \\
\;$+$SS          & \textbf{57.98} & 42.99 & 64.14 & 53.77 \\
\;$+$SS$+$RP     & n/a & \textbf{43.48} & 64.33 & 54.41 \\
\bottomrule
\end{tabular}
\caption{Best in-domain average reached at the $\kappa$-rule point (best configuration per method and suite). RP is inapplicable on T1 (no non-target residual).}

\label{tab:ablation}
\end{table}

\subsection{Component Effects in the Two Cases}
\label{sec:ablations}

Table~\ref{tab:ablation} evaluates each component under the $\kappa$-rule for CORAM in the top half and CORAM-C in the bottom half. CORAM-C performs better than plain CORAM on T2 by $+0.21$ and on T3 by $+0.07$. Combining spread slicing and the residual pathway gives the best value at the selected $\kappa$ on two of the four suites, with 43.48 on T2, and 54.71 on T4.  On T1, CORAM-C$+$SS gives the best result under the $\kappa$-rule at 57.98, while CORAM$+$RP gives the best result on T3 at 64.38. Sweeping $\lambda$ further improves the T1 result to 58.21 at $\lambda=2.30$ and the T4 result to 54.99 at $\lambda=1.25$.
These gains quantify the remaining gap between the $\kappa$-rule point and the swept optimum.

\paragraph{Effect of spread slicing.} The sweep results below are reported in the supplementary material. When the expert-update magnitudes are comparable, spread slicing improves the swept optimum by $+0.34$ on T1, $+0.41$ on T2, and $+0.10$ on T3 relative to the best contiguous configuration on the same branch. When the update magnitudes differ substantially, spread slicing alone does not help. Every spread-only configuration on T4 is $0.58$ to $0.86$ points below its contiguous counterpart. 

The reduction is concentrated in the safety task. Four of the five T4 domains change by less than $\pm1$ point, while safety decreases by $4.7$ points. The supplementary analysis links this drop to less stable open-ended generation rather than a general capability loss. Adding the residual pathway restores the omitted embedding and normalization updates. The combined configuration outperforms the residual pathway alone, with 54.62 vs.\ 54.56 at the selected $\kappa$ and 54.99 vs.\ 54.84 at the swept optimum in Table~\ref{tab:t4}. 
Spread slicing works by itself when update magnitudes are comparable, while uneven updates benefit from the residual pathway. 

\section{Limitations}
\label{sec:limitations}
\paragraph{Intermediate values of $D$.} Four bases have low $D$, while one has high $D$. We do not extrapolate to intermediate values, which are not represented in current public expert suites.

\paragraph{Scale and strength.} 

Proposition~\ref{prop:sqrtn} gives the ideal scale, and $c_{\mathrm{RMS}}$ extends it to correlated or uneven updates. It remains within $10\%$ of $\sqrt{N}$ when update magnitudes are comparable. However, $\kappa\approx1.15$ and $\kappa\approx0.5$ are empirical values selected by $D$. The latter is calibrated on a single high-$D$ base. A full contraction analysis of the composed slice map remains open.


\paragraph{Behavioral evidence.} 
The failure of spread slicing under uneven updates is observed at the domain and generation levels and is corrected by the residual pathway. We do not provide a theoretical explanation.

\paragraph{Scope.} We follow the expert suites and protocols of MergeBench and \citet{orthomerge}. Larger numbers of experts, other architectures such as mixture-of-experts models, and experts derived from different base checkpoints are not evaluated.

\section{Conclusion}
\label{sec:conclusion}

We presented CORAM, a slice-level method for merging finetuned experts.  
Each expert slice is represented by its SVD in the base frame, and the task-specific factors are averaged on their corresponding manifolds. 
CORAM compensates for the resulting contraction using 
an amplification scale estimated from the norms of the expert and merged updates and a strength selected from update dispersion without evaluating candidate merged models. 
Spread slicing helps when update magnitudes are comparable, while uneven updates benefit from the residual pathway. 
Across four suites covering three model families and both language and vision-language experts, CORAM improves over Orthogonal Model Merging and matches or exceeds the strongest weight-space baselines without a per-benchmark sweep over $\lambda$. 
Future work will study intermediate update dispersion and contraction under manifold averaging.

\bibliography{aaai2027}

\end{document}


\maketitle

This supplement provides the method details, complete experimental
results, and additional analyses omitted from the main paper.

\begin{itemize}
\item \textbf{Section~A}: the complete CORAM pipeline, sign
alignment, the numerical implementation of the geometric operators,
conflict-aware variants, spread slicing, and the residual pathway.
\item \textbf{Section~B}: the branch diagnosis that motivates
amplification, the proof of Proposition~1, and the measurements
behind the contraction scale and the restoration strength.
\item \textbf{Section~C}: the evaluation protocol, computing
infrastructure, baselines, and configuration naming, with the
mapping to the main tables.
\item \textbf{Section~D}: the retained configurations, coefficient
sweeps, the rule-versus-optimum comparison, the complete
configuration grid, threshold sensitivity, leave-one-base-out
validation, and alternative coefficient choices.
\item \textbf{Section~E}: component ablations and the additional
analysis of spread slicing on T4.
\item \textbf{Section~F}: the slice-height study.
\item \textbf{Section~G}: the OOD settings and the OrthoMerge
reproduction.
\item \textbf{Section~H}: multi-seed results and confidence
intervals.
\item \textbf{Section~I}: computational cost.
\end{itemize}

\section{A. Additional Method Details}
\label{app:method}

This section gives the implementation details omitted from the main
paper, including the complete merging pipeline, sign alignment,
conflict-aware variants, spread slicing, and the residual pathway.

\paragraph{Complete pipeline.}
For each target matrix, the base and expert weights are first divided
into row slices. Each expert slice is represented in the corresponding
base-model SVD frame by a rotation, a relative spectral shift, and a
right factor. The three factors are merged in their respective spaces,
and the merged slice is reconstructed in the base frame. The geometric
update is then amplified using the coefficient defined in the main
paper. Spread slicing, when enabled, is applied before factorization,
while the residual pathway is added after the geometric merge.

\paragraph{Base-frame factorization and sign alignment.}
For a base slice
\(
W_0^{(g)}=U_0S_0V_0^\top
\)
and expert slice
\(
W_i^{(g)}=U_iS_iV_i^\top,
\)
the singular-vector signs are aligned jointly against the base pair
\((U_0,V_0)\). Each pair \((u,v)\) is flipped together, so the expert
slice remains unchanged. The aligned factors define
\[
Q_i\in \mathrm{SO}(h),\qquad
dS_i=S_iS_0^{-1}-I,\qquad
V_i\in\mathrm{St}(d_{\mathrm{in}},h).
\]
The rotation, relative spectral shift, and right factor are then merged
as described in the main paper.

The rotation-side polar factor applies an explicit determinant
correction: from \(U,S,V^{\top}=\mathrm{SVD}(M)\) we form
\(Q=UV^{\top}\), and whenever \(\det Q<0\), the sign of the last column
of \(U\) is flipped and \(Q\) is recomputed, guaranteeing
\(Q\in\mathrm{SO}(h)\) rather than merely \(O(h)\). The right-factor
polar projection applies no such correction, since
\(V_i\in\mathrm{St}(d_{\mathrm{in}},h)\) is not square and the
determinant constraint does not apply there.

\paragraph{Matrix logarithm and exponential.}
The rotation logarithm is computed in double precision through the
Cayley transform: \(W=(Q+I)^{-1}(Q-I)\) is real skew-symmetric with
eigenvalues \(i\tan(\theta/2)\), so \(iW\) is Hermitian and a
GPU-native Hermitian eigendecomposition recovers the rotation angles;
the generator is reassembled from the eigenbasis, the real part is
taken, and an explicit skew-symmetric projection
\(K\leftarrow\tfrac{1}{2}(K-K^{\top})\) is applied. When \(Q\) has an
eigenvalue near \(-1\) (\(\theta\approx\pi\)), where the Cayley
transform is ill-conditioned, the implementation falls back to a
complex general eigendecomposition with the principal logarithm. The
two paths agree to about \(10^{-11}\). The exponential map applies the
skew projection, evaluates the matrix exponential, and re-projects the
result onto the orthogonal group by a polar decomposition.
Table~\ref{tab:numaudit} reports the measured reconstruction error
\(\|\exp(\log Q)-Q\|_F/\|Q\|_F\) over all slices of every suite.

\paragraph{Singular-value stabilization.}
The relative spectral shift divides by the base spectrum through a
truncated inverse, \(dS_i=S_i/\max(S_0,\epsilon)-1\) with
\(\epsilon=10^{-12}\); slice SVDs are computed in batched double
precision. Table~\ref{tab:numaudit} reports the distribution of the
slice condition numbers \(\sigma_{\max}/\sigma_{\min}\) of \(S_0\) for
contiguous and for variance-spread slicing. Spread slicing leaves the
bulk of the distribution unchanged and collapses its tail, consistent
with the conditioning motivation given in the main paper.

\begin{table*}[!tp]
\centering
\small
\setlength{\tabcolsep}{4pt}
\begin{tabular}{lcccccccc}
\toprule
& \multicolumn{3}{c}{Rotation log (all slices)}
& \multicolumn{3}{c}{Cond.\ (contiguous)}
& \multicolumn{2}{c}{Cond.\ (spread)}\\
Suite & Mean err & Max err & \(\theta{>}3\)
& Median & p95 & Max & p95 & Max\\
\midrule
T1 & \(8.8{\times}10^{-16}\) & \(1.9{\times}10^{-15}\) & 0 & 1.3 & 2.3 & 706 & 2.8 & 38\\
T2 & \(8.8{\times}10^{-16}\) & \(2.2{\times}10^{-15}\) & 0 & 1.3 & 2.6 & 822 & 3.1 & 28\\
T3 & \(8.8{\times}10^{-16}\) & \(1.9{\times}10^{-15}\) & 0 & 1.3 & 5.2 & 635 & 5.3 & 282\\
T4 & \(1.4{\times}10^{-15}\) & \(5.1{\times}10^{-10}\) & 64 & 1.3 & 2.0 & 43 & 2.5 & 18\\
\bottomrule
\end{tabular}
\caption{Numerical audit of the geometric operators at \(h=8\), over
every slice of every expert (0.48M--1.16M rotations per suite).
Rotation-logarithm reconstruction errors are at machine precision;
the near-\(\pi\) fallback fires on 64 of 1.16M rotations on T4 only
and is handled by the exact eigendecomposition path. No slice has
\(\sigma_{\min}<10^{-8}\). Variance-spread slicing leaves the median
condition number unchanged but collapses the tail (maximum
\(706\!\to\!38\), \(822\!\to\!28\), \(635\!\to\!282\),
\(43\!\to\!18\)), confirming its conditioning motivation.}
\label{tab:numaudit}
\end{table*}

\paragraph{Conflict-aware variants.}
For expert update
\(
\tau_i=W_i^{(g)}-W_0^{(g)}
\)
and mean update
\(
\bar{\tau}=\frac{1}{N}\sum_i\tau_i,
\)
column \(j\) of expert \(i\) is flagged when
\[
\cos\bigl(\tau_i[:,j],\bar{\tau}[:,j]\bigr)<0.
\]
The flagged and non-flagged columns are assigned either to the
geometric merge or to a Euclidean residual. We evaluate three
variants:

\begin{itemize}
    \item The non-flagged columns are merged geometrically, and the
    flagged columns are discarded.
    \item The non-flagged columns are merged geometrically, and the
    flagged columns are averaged and added back as a residual.
    \item The flagged columns are merged geometrically, and the
    non-flagged columns are averaged and added back as a residual.
\end{itemize}

The fourth combination, in which only the flagged columns are retained
and the non-flagged columns are discarded, removes the signal on which
the experts agree and is not used. Since masking changes the slice, the
masked target is factorized again. Plain CORAM applies no conflict mask
and reads the cached factors directly.

\begin{figure}[!tbp]
\centering
\IfFileExists{AuthorKit27/Figures/spreadslice.png}{
  \includegraphics[width=0.92\columnwidth]{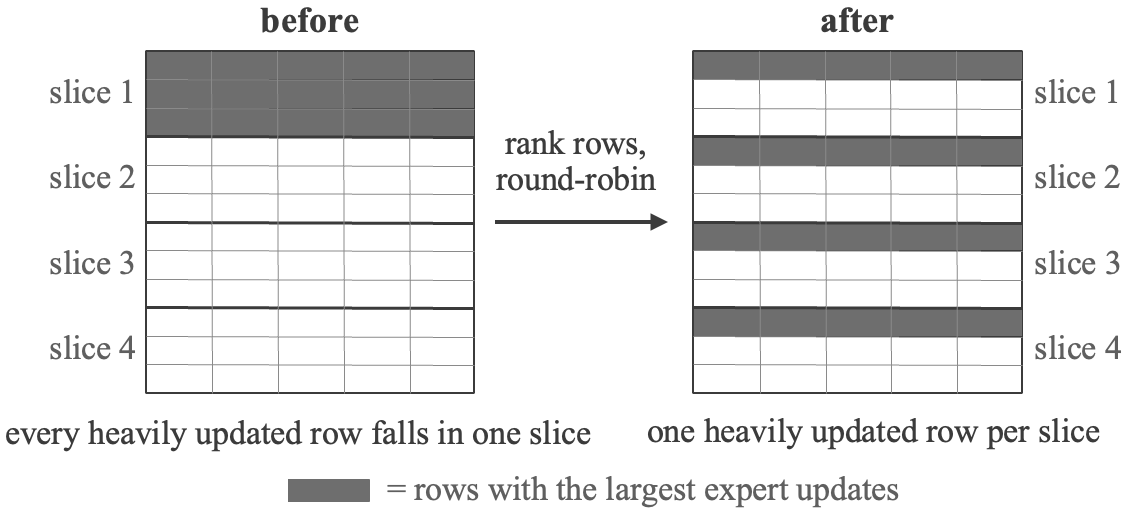}
}{
  \fbox{
    \parbox[c][1.25in][c]{0.84\columnwidth}{
      \centering
      \pending{\texttt{AuthorKit27/Figures/spreadslice.png} is uploaded on
      Overleaf; it is not present in this local tree.}
    }
  }
}
\caption{Spread slicing. Rows are scored by their relative update
magnitudes, ranked by the selected priority, and assigned to slices so
that highly updated rows are distributed evenly. The inverse
permutation restores the original row order after merging.}
\label{fig:spread_slicing}
\end{figure}

\paragraph{Spread slicing.}
Contiguous slicing follows the original row order. Spread slicing
instead assigns rows to slices according to their relative update
magnitudes, as illustrated in Figure~\ref{fig:spread_slicing}. For row
\(r\) and expert \(i\), we use
\[
s_i(r)=
\frac{\|\Delta W_i[r,:]\|}
     {\|W_0[r,:]\|}.
\]
Let \(s^{(1)}(r)\) and \(s^{(2)}(r)\) denote the largest and second
largest values across experts. We consider four priorities:

\begin{itemize}
    \item \textbf{Mean:}
    \(
    \frac{1}{N}\sum_i s_i(r).
    \)
    \item \textbf{Energy:}
    \(
    s^{(1)}(r).
    \)
    \item \textbf{Variance:}
    \(
    \mathrm{Var}_i[s_i(r)]\,s^{(1)}(r).
    \)
    \item \textbf{Owner balance:}
    \(
    \bigl(s^{(1)}(r)-s^{(2)}(r)\bigr)s^{(1)}(r).
    \)
\end{itemize}

Mean, energy, and variance priorities assign rows in round-robin order.
Owner balance assigns each row to the slice with the smallest
accumulated priority and uses the dominant expert to break ties. The
same permutation is applied to the base and all experts. After merging,
the inverse permutation restores the original row order.

\paragraph{Residual pathway.}
The geometric merge is applied to the target attention and MLP linear
layers. Embeddings, normalization parameters, and other non-target
parameters are merged separately. For non-target update
\(
\rho_i=W_i-W_0,
\)
we use task arithmetic, TIES, or Task Singular Vectors and add the
result as an unscaled residual patch:
\[
W^*
=
W_0+
\begin{cases}
\lambda\bigl(W_{\mathrm{merge}}-W_0\bigr),
& \text{target layers},\\[2mm]
\bar{\rho},
& \text{non-target layers}.
\end{cases}
\]
The amplification coefficient is applied only to the geometric update.
When spread slicing and the residual pathway are both enabled, the row
permutation is applied before the geometric merge and the residual
patch is added afterward.

\section{B. Amplitude Restoration}
\label{app:amplitude}

This section reports the branch diagnosis that motivates
amplification, provides the proof of the cancellation scale, relates
it to the three per-slice factors, and reports the measurements used
to set the restoration strength.

\paragraph{Branch diagnosis and the necessity of amplification.}
Before deriving the amplification rule, we verify empirically which
per-slice factors carry the task signal and how much magnitude the
geometric merge loses. On T2 (Llama-3.2-3B, \(h{=}16\), uniform
expert weights, no amplification, i.e.\ \(\lambda{=}1\)), we
reconstruct the merged model from every subset of the three factors,
keeping the remaining factors at their base values. Table~\ref{tab:branch_diag} reports the
in-domain average together with the magnitude and direction of the
merged update \(\Delta = W_{\mathrm{merge}} - W_0\) relative to the
unscaled task-arithmetic sum
\(\Delta_{\mathrm{TA}} = \sum_i (W_i - W_0)\).

\begin{table}[!tbp]
\centering
\small
\setlength{\tabcolsep}{5pt}
\begin{tabular}{lccc}
\toprule
Branch & Avg & \(\|\Delta\|/\|\Delta_{\mathrm{TA}}\|\)
& \(\cos(\Delta,\Delta_{\mathrm{TA}})\)\\
\midrule
\(dS\) only & 27.23 & 0.006 & 0.024\\
\(Q\) only & 26.97 & 0.768 & 0.001\\
\(V\) only & 33.98 & 0.788 & 0.253\\
\(Q{+}dS\) & 27.30 & 0.768 & 0.001\\
\(dS{+}V\) & 34.24 & 0.788 & 0.253\\
\(Q{+}V\) (\texttt{lr}) & 34.69 & 0.207 & 0.966\\
\(Q{+}dS{+}V\) (\texttt{full}) & 33.94 & 0.207 & 0.967\\
\midrule
Base model & 27.18 & -- & --\\
Task arithmetic & 38.27 & 1.000 & 1.000\\
\bottomrule
\end{tabular}
\caption{Branch diagnosis on T2 (Llama-3.2-3B, \(h{=}16\),
\(\lambda{=}1\)). Each row merges a subset of the per-slice factors
and keeps the rest at their base values. Only the branches containing
both rotations are direction-correct, and their magnitude contracts
to \(0.21\,\|\Delta_{\mathrm{TA}}\|\). The sweep optima in
Table~\ref{tab:optt2} are at \(h{=}8\).}
\label{tab:branch_diag}
\end{table}

Three observations follow. First, the spectral shift alone is inert:
its update norm is \(10^{-4}\) of \(\|W_0\|\) and its score matches
the base model. Second, the left rotation alone produces a sizeable
but task-orthogonal perturbation (\(\cos\approx0\)) and scores below
the base model. The right factor alone recovers only part of the
signal and collapses onto the mathematics expert, which dominates 193
of the 196 target modules. Only the two branches that merge both
rotations point in the task direction, with
\(\cos(\Delta,\Delta_{\mathrm{TA}})\approx0.97\). These are the
\texttt{lr} and \texttt{full} lines retained in the configuration
tables. Third, for these two branches the update magnitude contracts
to \(0.207\,\|\Delta_{\mathrm{TA}}\|\), because the two
rotation-induced updates largely cancel:
\(\|\Delta_{Q+V}\|\,/\,(\|\Delta_{Q}\|^{2}+\|\Delta_{V}\|^{2})^{1/2}
= 0.188\). In this diagnostic, the merge is well aligned with the
task-arithmetic direction but substantially under-scaled. This
observation motivates the cancellation analysis below. Restoring the
magnitude recovers the performance: the same \texttt{full} line,
swept over \(\lambda\) at \(h{=}8\), reaches \(42.76\) at
\(\lambda^{*}=2.80\)
(Table~\ref{tab:optt2}), compared with \(33.94\) at \(\lambda{=}1\).
Amplification is therefore a necessary part of the method, not a
tuning refinement.

\paragraph{Proof of Proposition 1.}
Let \(\tau_1,\dots,\tau_N\) satisfy
\(
\mathbb{E}\langle\tau_i,\tau_j\rangle=0
\)
for \(i\neq j\) and
\(
\|\tau_i\|=\|\tau\|
\)
for all \(i\). Expanding the squared norm of the mean gives
\begin{align}
\mathbb{E}\left\|
\frac{1}{N}\sum_i\tau_i
\right\|^2
&=
\frac{1}{N^2}
\left(
\sum_i\mathbb{E}\|\tau_i\|^2
+
\sum_{i\neq j}
\mathbb{E}\langle\tau_i,\tau_j\rangle
\right)
\nonumber\\
&=
\frac{1}{N^2}\,N\|\tau\|^2
=
\frac{\|\tau\|^2}{N}.
\end{align}
The mean therefore has RMS norm
\(
\|\tau\|/\sqrt{N},
\)
and restoring it to the typical expert magnitude requires
amplification by \(\sqrt{N}\). \hfill\(\square\)

\paragraph{Unequal norms.}
The equal-norm assumption can be dropped without changing the scale.
If the updates remain incoherent,
\(\mathbb{E}\langle\tau_i,\tau_j\rangle=0\) for \(i\neq j\), but the
norms \(\|\tau_i\|\) differ, the same expansion gives
\[
\mathbb{E}\left\|\frac{1}{N}\sum_i\tau_i\right\|^2
=\frac{1}{N^2}\sum_i\|\tau_i\|^2 .
\]
Restoring the mean to the RMS expert magnitude therefore requires
\[
\frac{\sqrt{\tfrac{1}{N}\sum_i\|\tau_i\|^2}}
     {\sqrt{\tfrac{1}{N^2}\sum_i\|\tau_i\|^2}}
=\sqrt{N},
\]
independently of how uneven the norms are. Norm imbalance alone does
not change the cancellation scale. The smaller optimal restoration
strength at high dispersion must therefore come from correlated
updates and from the nonlinear parts of the slice map, not from the
norm imbalance itself. \(D\) should accordingly be read as an
empirical predictor of the high-dispersion case: it is a proxy for
the correlation and nonlinear effects that reduce the useful
restoration strength, not their direct cause.

\paragraph{Relation to the per-slice factors.}
The cancellation result applies directly to the two factors averaged
in linear spaces. The rotation generators are averaged in
\(\mathfrak{so}(h)\), and the relative spectral shifts are averaged in
\(\mathbb{R}^{h}\). The right factors are instead averaged by a
Euclidean sum followed by polar projection. Proposition~1 therefore
motivates the \(\sqrt{N}\) scale for the first two factors, while
\(c_{\mathrm{RMS}}\) measures the contraction of the complete
reconstructed update:
\[
c_{\mathrm{RMS}}
=
\frac{\mathrm{RMS}_i\|\Delta W_i\|}
     {\|\Delta W_{\mathrm{merge}}\|}.
\]

\begin{table}[!tbp]
\centering
\small
\setlength{\tabcolsep}{4pt}
\begin{tabular}{lccccc}
\toprule
Base & \(N\) & \(D\) & \(c_{\mathrm{RMS}}\) & \(\sqrt{N}\) & Rel.\ diff.\\
\midrule
Llama-3.1-8B
& 5
& 3.5
& 2.09
& 2.24
& 6.7\%\\
Llama-3.2-3B
& 5
& 2.6
& 2.03
& 2.24
& 9.4\%\\
Qwen2.5-VL-7B
& 3
& 1.3
& 1.72
& 1.73
& 0.6\%\\
Gemma-2-2B
& 5
& 1.67
& 2.09
& 2.24
& 6.6\%\\
Gemma-2-9B
& 5
& 16.0
& 2.19
& 2.24
& 2.3\%\\
\bottomrule
\end{tabular}
\caption{Measured contraction scale. The first four bases form the
low-dispersion case. The Gemma-2-9B experts have substantially more
uneven update magnitudes.}
\label{tab:crms}
\end{table}

\paragraph{Measured contraction scale.}
Table~\ref{tab:crms} compares \(c_{\mathrm{RMS}}\) with \(\sqrt{N}\).
The approximation is close for the four low-dispersion settings. The
Gemma-2-9B setting has substantially more uneven expert-update
magnitudes and is treated separately when selecting the restoration
strength.

\paragraph{Restoration strength.}
The scale \(c_{\mathrm{RMS}}\), or its default approximation
\(\sqrt{N}\), determines the size of the contraction correction. The
factor \(\kappa\) determines how much of this scale is restored. The
swept optima form two cases. For the four bases with
\(D\in[1.3,3.5]\), the useful values of \(\kappa\) are approximately
\(1.0\)--\(1.2\). For Gemma-2-9B, where \(D\approx16\), the useful
value is near \(0.5\). We use the shared constants
\[
\kappa=1.15
\qquad\text{and}\qquad
\kappa=0.5
\]
for the low- and high-dispersion cases, respectively.

\begin{figure}[!tbp]
\centering
\IfFileExists{AuthorKit27/Figures/fig2_dichotomy.png}{
  \includegraphics[width=0.92\columnwidth]{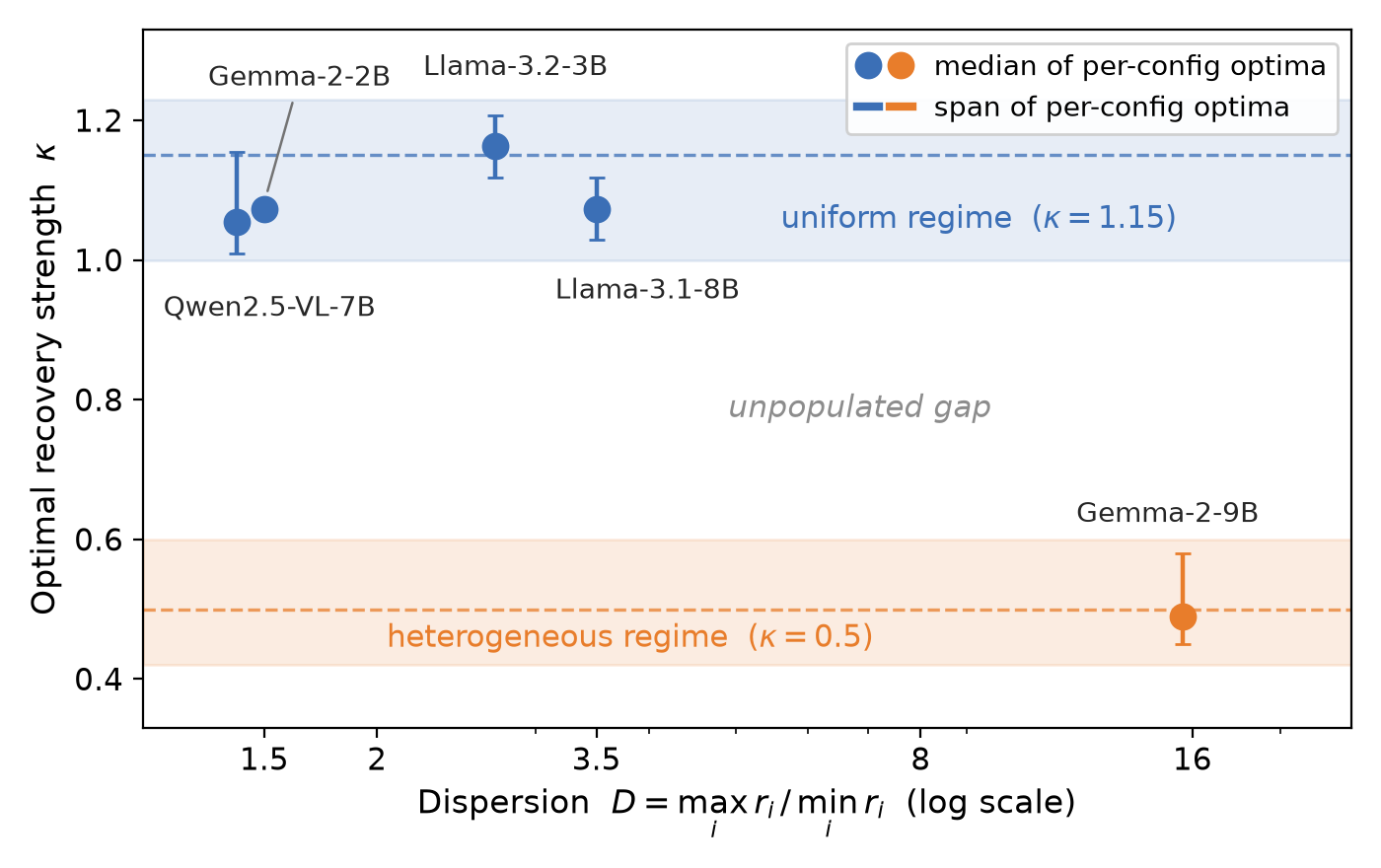}
}{
  \fbox{
    \parbox[c][1.35in][c]{0.84\columnwidth}{
      \centering
      \pending{Insert the plot of swept $\kappa_{\mathrm{opt}}$ against
      $D$ for the five bases.}
    }
  }
}
\caption{Swept restoration strength
\(\kappa_{\mathrm{opt}}=\lambda^*/\sqrt{N}\) against the dispersion statistic \(D\). Dashed lines mark the two shared constants.}
\label{fig:kappa_cases}
\end{figure}

Figure~\ref{fig:kappa_cases} plots the swept
\(
\kappa_{\mathrm{opt}}=\lambda^*/\sqrt{N}
\)
against \(D\). The available bases form two separated groups. The
current expert suites do not cover intermediate values of \(D\), and
we do not define an interpolation between the two constants.

\section{C. Experimental Protocol and Configuration Definitions}
\label{app:setup}

This section specifies the evaluation settings and the configuration
names used in the supplementary tables.

\subsection{Evaluation Protocol}

\paragraph{T1: Llama-3.1-8B.}
T1 merges five orthogonally finetuned experts. The in-domain tasks are
MATH500, HumanEval+, ScienceQA, CommonsenseQA, and Social-IQA. The
out-of-domain (OOD) tasks are M-ARC and AGIEval. All OOD evaluations
use the same zero-shot setting across methods.

\paragraph{T2 and T4: MergeBench.}
T2 and T4 merge five fully finetuned experts based on Llama-3.2-3B and
Gemma-2-9B, respectively. The five domains are instruction following,
mathematics, coding, multilingual understanding, and safety.

Instruction following is evaluated using IFEval strict accuracy.
Mathematics uses GSM8K with 8-shot chain-of-thought prompting and
strict matching. Coding is the mean of HumanEval+ and MBPP+ pass@1.
Multilingual performance is averaged over the MergeBench multilingual
tasks. Safety is computed as
\[
\mathrm{Safety}
=
\mathrm{mean}
\left(
1-\mathrm{harm},
1-\mathrm{ASR},
1-\mathrm{ASR}_{\mathrm{jb}},
\mathrm{acc}_{\mathrm{or}}
\right).
\]
The OOD tasks are MMLU\(^{\dagger}\), with mathematics and coding
subsets removed, and AGIEval.

\paragraph{T3: Qwen2.5-VL-7B-Instruct.}
T3 merges three vision--language experts. The evaluation tasks are
MMSI-Bench, EmbSpatial, the multiple-choice subset of MMMU-Med,
PathVQA, OCRBench, and CharXiv. CharXiv uses its official GPT judge
with \texttt{gpt-4o-2024-05-13}, the official prompt, and the official
parsing procedure. IFEval and MMBench are used for the OOD columns.

\paragraph{Evaluation harnesses.}
Language evaluations use a fixed lm-evaluation-harness configuration.
Code tasks use the bigcode evaluation harness. Vision--language tasks
use lmms-eval. Safety follows the MergeBench safety stack.

Merging is deterministic. Unless otherwise stated, each reported score
is a single evaluation run; the multi-seed study in Section~H reports
variation across decoding seeds for the main-table checkpoints.

\paragraph{Computing infrastructure.}
Experiments ran on four Linux nodes: one with NVIDIA L40S GPUs (46~GB
each; AMD EPYC 7513, 512~GB RAM, Ubuntu 22.04, CUDA 13.0) and three
with eight NVIDIA B300 GPUs each (275~GB each; Intel Xeon 6737P, 3~TB
RAM, Ubuntu 24.04, CUDA 13.2). Merging uses PyTorch 2.7/2.8 with
safetensors. Language evaluations use lm-evaluation-harness 0.4.9.1
with transformers 4.57.6 and datasets 3.1.0. Code evaluations use the
bigcode evaluation harness with transformers 5.12.1. Vision--language
evaluations use lmms-eval 0.5.0 with qwen-vl-utils 0.0.14. Safety
evaluations use vLLM 0.9.0; Gemma-2-9B safety generation instead uses
the HF backend with the Gemma chat template (see Section~H). Decoding
is greedy unless a task specifies otherwise; coding tasks use
temperature 0.2 with \(n=10\). Exact harness commits will be included
with the released code.

\subsection{Baselines}

Linear averaging, task arithmetic, TIES, and DARE-TIES are implemented
as per-tensor algorithms following their original definitions. Their
coefficient settings follow the MergeBench protocol. For OrthoMerge,
we evaluate all released variants under the same harness and report
the strongest result for each suite.

\subsection{Configuration Naming}

Table~\ref{tab:config_names} defines the internal configuration names
used in the complete result tables.

\begin{table*}[!tp]
\centering
\small
\setlength{\tabcolsep}{6pt}
\renewcommand{\arraystretch}{1.08}
\begin{tabular}{lp{0.78\textwidth}}
\toprule
Name & Description\\
\midrule

\texttt{full}
&
Merges the rotation, relative spectral shift, and right factor. No
conflict masking, spread slicing, or residual pathway is applied.
\\

\texttt{lr}
&
Merges the rotation and right factor while keeping the spectral factor
at its base value \(S_0\).
\\

\texttt{nca}
&
Merges the non-conflicting columns geometrically and discards the
conflicting columns.
\\

\texttt{nca+ave}
&
Merges the non-conflicting columns geometrically and adds the average
of the conflicting columns as a Euclidean residual.
\\

\texttt{ca}
&
Merges the conflicting columns geometrically and discards the
non-conflicting columns.
\\

\texttt{ca+ave}
&
Merges the conflicting columns geometrically and adds the average of
the non-conflicting columns as a Euclidean residual.
\\

\texttt{line.ta}
&
Uses the geometric configuration specified by \texttt{line} for the
target layers and task arithmetic for the non-target residual pathway.
\\

\texttt{line.ties}
&
Uses the geometric configuration specified by \texttt{line} for the
target layers and TIES for the non-target residual pathway.
\\

\texttt{line.tsvm}
&
Uses the geometric configuration specified by \texttt{line} for the
target layers and Task Singular Vectors for the non-target residual
pathway.
\\

\texttt{SS(line,mode)}
&
Applies spread slicing to the configuration specified by
\texttt{line}. The priority mode is
\texttt{mean}, \texttt{energy}, \texttt{variance}, or
\texttt{owner}; \texttt{owner} denotes the owner-balance priority
defined in the main paper.
\\

\bottomrule
\end{tabular}
\caption{Configuration names used in the supplementary tables. The
suffixes \texttt{ta}, \texttt{ties}, and \texttt{tsvm} denote the
merger used for the non-target residual pathway.}
\label{tab:config_names}
\end{table*}
The residual pathway is not applicable to T1 because the released OFT
experts modify only the target linear layers.

\subsection{Mapping to the Main Tables}

Table~\ref{tab:main_mapping} records the exact supplementary
configuration used for each CORAM row in the main paper.

\begin{table*}[!tp]
\centering
\scriptsize
\setlength{\tabcolsep}{5pt}
\begin{tabular}{lllllllll}
\toprule
Suite
& Main-paper row
& Internal line
& \(h\)
& SS priority
& RP merger
& Conflict variant
& Rule \(\lambda\)
& Best \(\lambda\)\\
\midrule
T1 & CORAM
& \texttt{full}
& 8
& None
& None
& None
& 2.60
& 2.40 (\(h{=}16\))\\
T1 & CORAM-C
& \texttt{nca+ave}
& 8
& None
& None
& nca+ave
& 2.60
& 2.25\\
T1 & CORAM-C+SS
& \texttt{SS(ca+ave,energy)}
& 8
& energy
& None
& ca+ave
& 2.60
& 2.30\\
\midrule
T2 & CORAM+SS+RP
& \texttt{SS(lr,mean).ties}
& 8
& mean
& TIES
& None
& 2.60
& 2.60\\
T2 & CORAM-C+RP
& \texttt{nca.ties}
& 8
& None
& TIES
& nca
& 2.60
& 2.60\\
T2 & CORAM-C+SS+RP
& \texttt{SS(ca+ave,variance).ties}
& 8
& variance
& TIES
& ca+ave
& 2.60
& 2.60\\
\midrule
T3 & CORAM
& \texttt{lr}
& 8
& None
& None
& None
& 2.00
& 1.60\\
T3 & CORAM+RP
& \texttt{lr.ta}
& 8
& None
& TA
& None
& 2.00
& 2.00\\
T3 & CORAM-C+RP
& \texttt{nca.ta}
& 8
& None
& TA
& nca
& 2.00
& 1.75\\
\midrule
T4 & CORAM-C
& \texttt{ca+ave}
& 8
& None
& None
& ca+ave
& 1.10
& 1.10\\
T4 & CORAM+RP
& \texttt{lr.ties}
& 8
& None
& TIES
& None
& 1.10
& 1.25\\
T4 & CORAM+SS+RP
& \texttt{SS(lr,mean).ta}
& 8
& mean
& TA
& None
& 1.10
& 1.25\\
\bottomrule
\end{tabular}
\caption{Mapping between the main-paper rows and the internal
configuration names (Table~\ref{tab:config_names}), with the
\(\kappa\)-rule and swept-best coefficients. All checkpoints use
\(h=8\), except the T1 CORAM best-\(\lambda\) entry (\(h=16\)).}
\label{tab:main_mapping}
\end{table*}

\section{D. Complete Configuration Results and Coefficient Sweeps}
\label{app:full_results}

This section reports the retained CORAM configurations. For each line,
the tables give the swept-optimal coefficient and the corresponding
in-domain average. For T3, the configuration and coefficient sweeps
in this section use the five deterministic non-judged tasks and
exclude CharXiv, because each CharXiv pass requires external GPT
judging. Main-table summaries and final checkpoint comparisons
include CharXiv and use the six-task average unless stated otherwise.
The complete machine-readable sweep logs will be included with the
released artifacts.

\subsection{Per-Suite Sweep Optima}

Tables~\ref{tab:optt1}--\ref{tab:optt4} report the per-line sweep
optima for the four suites. Each table lists, for the retained lines
of one suite, the swept-optimal coefficient and the in-domain average
at that coefficient.

\begin{table}[!tbp]
\centering
\footnotesize
\setlength{\tabcolsep}{2.5pt}
\begin{tabular}{lcc|lcc}
\toprule
Line & \(\lambda^{*}\) & Avg
& Line & \(\lambda^{*}\) & Avg\\
\midrule
full
& 2.50 & 57.94
& SS(full,mean)
& 2.40 & 58.08\\
lr
& 2.20 & 57.80
& SS(full,variance)
& 2.30 & 57.99\\
nca
& 2.50 & 57.73
& SS(full,energy)
& 2.40 & 57.97\\
nca+ave
& 2.25 & 57.74
& SS(ca+ave,energy)
& 2.30 & \textbf{58.21}\\
ca
& 2.10 & 47.45
& full (\(h=16\))
& 2.40 & 57.98\\
ca+ave
& 2.50 & 57.87
& lr (\(h=16\))
& 2.40 & 57.97\\
\bottomrule
\end{tabular}
\caption{T1 per-line sweep optima. Strict conflict masking without an
averaged residual gives a substantially lower result on T1.}
\label{tab:optt1}
\end{table}

\begin{table}[!tbp]
\centering
\footnotesize
\setlength{\tabcolsep}{2.5pt}
\begin{tabular}{lcc|lcc}
\toprule
Line & \(\lambda^{*}\) & Avg
& Line & \(\lambda^{*}\) & Avg\\
\midrule
full
& 2.80 & 42.76
& nca.ta
& 2.60 & 43.00\\
lr
& 2.75 & 42.44
& nca.ties
& 2.60 & 43.32\\
nca
& 2.50 & 42.96
& nca.tsvm
& 2.80 & 43.06\\
nca+ave
& 2.60 & 42.33
& nca+ave.ties
& 2.55 & 43.18\\
ca+ave
& 2.60 & 42.56
& ca+ave.ties
& 2.55 & 42.96\\
full.ties
& 2.60 & 43.03
& SS(lr,mean)
& 2.80 & 42.99\\
lr.ties
& 2.60 & 43.11
& SS(ca+ave,variance).ties
& 2.60 & \textbf{43.48}\\
\bottomrule
\end{tabular}
\caption{Selected T2 per-line sweep optima at \(h=8\).}
\label{tab:optt2}
\end{table}

\begin{table}[!tbp]
\centering
\footnotesize
\setlength{\tabcolsep}{2.5pt}
\begin{tabular}{lcc|lcc}
\toprule
Line & \(\lambda^{*}\) & Avg
& Line & \(\lambda^{*}\) & Avg\\
\midrule
full
& 1.60 & 63.25
& nca.ta
& 1.75 & 63.75\\
lr
& 1.60 & 63.43
& nca+ave.ta
& 1.70 & 63.60\\
nca
& 1.60 & 63.07
& ca+ave.ta
& 1.70 & 63.40\\
nca+ave
& 1.75 & 63.42
& lr.ta
& 2.00 & 63.64\\
ca+ave
& 1.75 & 63.41
& SS(nca,energy)
& 1.60 & 63.72\\
full.ta
& 1.80 & 63.61
& SS(nca+ave,variance)
& 1.75 & 63.73\\
lr.ties
& 1.50 & 63.10
& SS(lr,owner).ta
& 2.00 & 63.66\\
\bottomrule
\end{tabular}
\caption{Selected T3 per-line sweep optima at \(h=8\). All entries are
five-task in-domain averages (CharXiv excluded).}
\label{tab:optt3}
\end{table}

\begin{table}[!tbp]
\centering
\footnotesize
\setlength{\tabcolsep}{2.5pt}
\begin{tabular}{lcc|lcc}
\toprule
Line & \(\lambda^{*}\) & Avg
& Line & \(\lambda^{*}\) & Avg\\
\midrule
full
& 1.20 & 54.56
& nca.ties
& 1.30 & 53.86\\
lr
& 1.00 & 54.74
& nca+ave.ties
& 1.20 & 54.72\\
nca
& 1.30 & 53.89
& ca+ave.ties
& 1.00 & 54.67\\
nca+ave
& 1.00 & 54.78
& lr.ties
& 1.25 & 54.84\\
ca+ave
& 1.10 & 54.32
& full.ties
& 1.20 & 54.67\\
lr.ta
& 1.10 & 54.48
& SS(lr,mean).ta
& 1.25 & \textbf{54.99}\\
\bottomrule
\end{tabular}
\caption{Selected T4 per-line sweep optima at \(h=8\). The optima lie
between \(\lambda=1.00\) and \(1.30\).}
\label{tab:optt4}
\end{table}

\subsection{Sweep Curves}

\begin{figure*}[!tp]
\centering
\IfFileExists{AuthorKit27/Figures/fig_lambda_sweeps.png}{
  \includegraphics[width=0.96\textwidth]{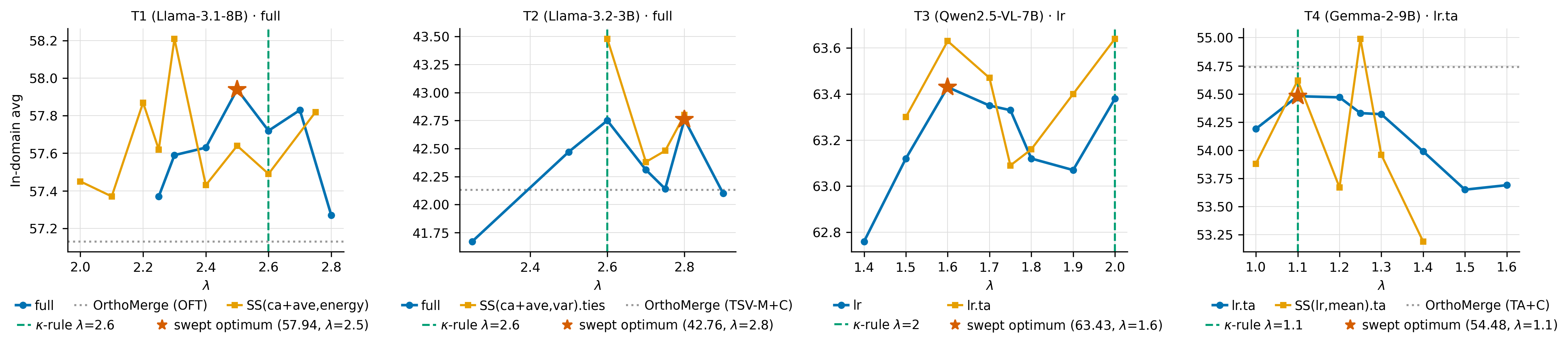}
}{
  \fbox{\parbox[c][1.3in][c]{0.9\textwidth}{\centering
  \pending{Place \texttt{AuthorKit27/Figures/fig\_lambda\_sweeps.png}.}}}
}
\caption{In-domain average versus \(\lambda\) at \(h=8\). Blue: the
representative line (\texttt{full} on T1/T2, \texttt{lr} on T3,
\texttt{lr.ta} on T4); orange: the strongest branch of the same
suite (SS(ca$+$ave,energy), SS(ca$+$ave,variance).ties, \texttt{lr.ta},
SS(lr,mean).ta). T3 curves are five-task averages, hence no
OrthoMerge reference line. Dashed vertical lines mark the
\(\kappa\)-rule coefficient; stars mark the swept optimum of the
representative line; dotted horizontal lines mark the strongest
OrthoMerge variant. The T4 panel is truncated at \(\lambda\le1.7\)
for readability; beyond this point the curve declines monotonically
to 43.5 at \(\lambda=2.9\) (full grid in
Table~\ref{tab:rule_vs_opt}).}
\label{fig:lambda_sweeps}
\end{figure*}

Figure~\ref{fig:lambda_sweeps} plots the in-domain average against
\(\lambda\) for two lines per suite: a representative plain line and
the strongest branch. The \(\kappa\)-rule coefficient, the swept
optimum of the representative line, and the strongest OrthoMerge
variant are marked. The curves are flat around their optima, so the
rule-selected coefficient loses at most a fraction of a point.

\subsection{Complete Rule-versus-Optimum Comparison}

The \(0.72\)-point bound stated in the main paper applies to the
configurations reported in the main tables, and every main-table
configuration satisfies it. Table~\ref{tab:rule_vs_opt} extends the
comparison to the broader exploratory grid: for every retained line,
it reports the score at the rule-selected coefficient and at the
swept optimum. Of the 52 rule-evaluated lines, 50 also satisfy the
bound. The two exceptions are exploratory combinations that were not
selected for any main-table row: \texttt{lr.ties} on T3 (\(0.89\);
the residual merger selected for T3 is task arithmetic, not TIES)
and \texttt{nca$+$ave} on T4 (\(0.73\)).

\begin{table*}[!tp]
\centering
\footnotesize
\setlength{\tabcolsep}{4pt}
\begin{tabular}{llccccc}
\toprule
Suite & Line & Rule \(\lambda\) & Rule score & \(\lambda^{*}\)
& Best score & Gap\\
\midrule
T1 & \texttt{full} & 2.60 & 57.72 & 2.50 & 57.94 & 0.22\\
 & \texttt{SS(full,mean)} & 2.60 & 57.49 & 2.40 & 58.08 & 0.59\\
 & \texttt{lr} & 2.60 & 57.48 & 2.20 & 57.80 & 0.32\\
 & \texttt{SS(full,variance)} & 2.60 & 57.47 & 2.30 & 57.99 & 0.52\\
 & \texttt{nca} & 2.60 & 57.54 & 2.50 & 57.73 & 0.19\\
 & \texttt{SS(full,energy)} & 2.60 & 57.72 & 2.40 & 57.97 & 0.25\\
 & \texttt{nca$+$ave} & 2.60 & 57.10 & 2.25 & 57.74 & 0.64\\
 & \texttt{SS(ca$+$ave,energy)} & 2.60 & 57.49 & 2.30 & 58.21 & 0.72\\
 & \texttt{ca} & 2.60 & 47.09 & 2.10 & 47.45 & 0.36\\
 & \texttt{full (h=16)} & 2.60 & 57.65 & 2.40 & 57.98 & 0.33\\
 & \texttt{ca$+$ave} & 2.60 & 57.60 & 2.50 & 57.87 & 0.27\\
 & \texttt{lr (h=16)} & 2.60 & 57.53 & 2.40 & 57.97 & 0.44\\
T2 & \texttt{full} & 2.60 & 42.75 & 2.80 & 42.76 & 0.01\\
 & \texttt{nca.ta} & 2.60 & 43.00 & 2.60 & 43.00 & 0.00\\
 & \texttt{lr} & 2.60 & 42.30 & 2.75 & 42.44 & 0.14\\
 & \texttt{nca.ties} & 2.60 & 43.32 & 2.60 & 43.32 & 0.00\\
 & \texttt{nca} & 2.60 & 42.96 & 2.50 & 42.96 & 0.00\\
 & \texttt{nca.tsvm} & 2.60 & 42.94 & 2.80 & 43.06 & 0.12\\
 & \texttt{nca$+$ave} & 2.60 & 42.33 & 2.60 & 42.33 & 0.00\\
 & \texttt{nca$+$ave.ties} & 2.60 & 42.69 & 2.55 & 43.18 & 0.49\\
 & \texttt{ca$+$ave} & 2.60 & 42.56 & 2.60 & 42.56 & 0.00\\
 & \texttt{ca$+$ave.ties} & 2.60 & 42.96 & 2.55 & 42.96 & 0.00\\
 & \texttt{full.ties} & 2.60 & 43.03 & 2.60 & 43.03 & 0.00\\
 & \texttt{SS(lr,mean)} & 2.60 & 42.93 & 2.80 & 42.99 & 0.06\\
 & \texttt{lr.ties} & 2.60 & 43.11 & 2.60 & 43.11 & 0.00\\
 & \texttt{SS(ca$+$ave,variance).ties} & 2.60 & 43.48 & 2.60 & 43.48 & 0.00\\
T3 & \texttt{full} & 2.00 & 63.18 & 1.60 & 63.25 & 0.07\\
 & \texttt{nca.ta} & 2.00 & 63.46 & 1.75 & 63.75 & 0.29\\
 & \texttt{lr} & 2.00 & 63.38 & 1.60 & 63.43 & 0.05\\
 & \texttt{nca$+$ave.ta} & 2.00 & 63.28 & 1.70 & 63.60 & 0.32\\
 & \texttt{nca} & 2.00 & 62.85 & 1.60 & 63.07 & 0.22\\
 & \texttt{ca$+$ave.ta} & 2.00 & 63.40 & 1.70 & 63.40 & 0.00\\
 & \texttt{nca$+$ave} & 2.00 & 63.37 & 1.75 & 63.42 & 0.05\\
 & \texttt{lr.ta} & 2.00 & 63.64 & 2.00 & 63.64 & 0.00\\
 & \texttt{ca$+$ave} & 2.00 & 63.20 & 1.75 & 63.41 & 0.21\\
 & \texttt{SS(nca,energy)} & 2.00 & 63.12 & 1.60 & 63.72 & 0.60\\
 & \texttt{full.ta} & 2.00 & 63.46 & 1.80 & 63.61 & 0.15\\
 & \texttt{SS(nca$+$ave,variance)} & 2.00 & 63.17 & 1.75 & 63.73 & 0.56\\
 & \texttt{lr.ties} & 2.00 & 62.21 & 1.50 & 63.10 & 0.89\\
 & \texttt{SS(lr,owner).ta} & 2.00 & 63.66 & 2.00 & 63.66 & 0.00\\
T4 & \texttt{full} & 1.10 & 54.33 & 1.20 & 54.56 & 0.23\\
 & \texttt{nca.ties} & 1.10 & 53.57 & 1.30 & 53.86 & 0.29\\
 & \texttt{lr} & 1.10 & 54.49 & 1.00 & 54.74 & 0.25\\
 & \texttt{nca$+$ave.ties} & 1.10 & 54.37 & 1.20 & 54.72 & 0.35\\
 & \texttt{nca} & 1.10 & 53.50 & 1.30 & 53.89 & 0.39\\
 & \texttt{ca$+$ave.ties} & 1.10 & 54.24 & 1.00 & 54.67 & 0.43\\
 & \texttt{nca$+$ave} & 1.10 & 54.05 & 1.00 & 54.78 & 0.73\\
 & \texttt{lr.ties} & 1.10 & 54.56 & 1.25 & 54.84 & 0.28\\
 & \texttt{ca$+$ave} & 1.10 & 54.32 & 1.10 & 54.32 & 0.00\\
 & \texttt{full.ties} & 1.10 & 54.29 & 1.20 & 54.67 & 0.38\\
 & \texttt{lr.ta} & 1.10 & 54.48 & 1.10 & 54.48 & 0.00\\
 & \texttt{SS(lr,mean).ta} & 1.10 & 54.62 & 1.25 & 54.99 & 0.37\\
\bottomrule
\end{tabular}
\caption{Rule-selected versus swept-optimal scores for every retained
line at \(h=8\) (plus the two retained \(h=16\) lines on T1). T3
entries are five-task averages. Lines whose swept optimum coincides
with the rule coefficient have zero gap by construction.}
\label{tab:rule_vs_opt}
\end{table*}

\subsection{The Complete Configuration Grid}

Tables~\ref{tab:grid_conflict}--\ref{tab:grid_ssrp_t4} list every
configuration line evaluated during the development of CORAM, grouped
by component layer: conflict variants under contiguous slicing,
spread-only lines, residual-pathway lines, and combined
spread$+$residual lines. Across this grid
(Tables~\ref{tab:grid_conflict}--\ref{tab:grid_ssrp_t4}), 238 lines
and 1{,}645 \((\text{line},\lambda)\) points were evaluated; the
slice-height study of Section~F is counted separately. Each row reports the
number of evaluated coefficients, their range, the swept optimum, and
the score at the rule-selected coefficient. Rule points absent from
the original sweeps were evaluated afterwards at the rule coefficient
(the T4 point with the main-table safety protocol), so every line
includes its rule-point entry. All lines use \(h{=}8\) (the \(h{=}16\) lines appear
in Table~\ref{tab:h16_existing}); T3 entries are five-task averages;
T4 entries include only evaluations passing the completeness checks.
Combinations not listed were not evaluated.

\begin{table}[!tbp]
\centering
\footnotesize
\setlength{\tabcolsep}{3pt}
\begin{tabular}{llccccc}
\toprule
Suite & Line & \(n_\lambda\) & \(\lambda\) range & \(\lambda^{*}\) & Best & Rule\\
\midrule
T1 & \texttt{ca} & 9 & 2.00--2.75 & 2.10 & 47.45 & 47.09\\
 & \texttt{ca$+$ave} & 12 & 1.00--3.25 & 2.50 & 57.87 & 57.60\\
 & \texttt{full} & 7 & 2.25--2.80 & 2.50 & 57.94 & 57.72\\
 & \texttt{lr} & 7 & 2.25--2.80 & 2.25 & 57.80 & 57.48\\
 & \texttt{nca} & 9 & 2.00--2.75 & 2.50 & 57.73 & 57.54\\
 & \texttt{nca$+$ave} & 12 & 1.00--3.25 & 2.25 & 57.74 & 57.10\\
T2 & \texttt{ca} & 7 & 2.50--3.00 & 2.80 & 26.66 & 26.37\\
 & \texttt{ca$+$ave} & 12 & 1.00--3.00 & 2.70 & 42.57 & 42.56\\
 & \texttt{full} & 8 & 2.00--2.90 & 2.80 & 42.76 & 42.75\\
 & \texttt{lr} & 8 & 2.00--2.90 & 2.75 & 42.44 & 42.30\\
 & \texttt{nca} & 9 & 2.30--3.00 & 2.50 & 42.96 & 42.96\\
 & \texttt{nca$+$ave} & 12 & 1.00--3.00 & 2.75 & 42.45 & 42.33\\
T3 & \texttt{ca} & 7 & 1.50--2.00 & 1.75 & 61.20 & 61.08\\
 & \texttt{ca$+$ave} & 9 & 1.00--2.00 & 1.75 & 63.41 & 63.20\\
 & \texttt{full} & 8 & 1.40--2.00 & 1.60 & 63.25 & 63.18\\
 & \texttt{lr} & 8 & 1.40--2.00 & 1.60 & 63.43 & 63.38\\
 & \texttt{nca} & 7 & 1.50--2.00 & 1.60 & 63.07 & 62.85\\
 & \texttt{nca$+$ave} & 9 & 1.00--2.00 & 1.75 & 63.42 & 63.37\\
T4 & \texttt{ca$+$ave} & 14 & 1.00--2.80 & 1.10 & 54.32 & 54.32\\
 & \texttt{full} & 14 & 0.80--2.90 & 1.20 & 54.56 & 54.33\\
 & \texttt{lr} & 17 & 0.80--2.90 & 1.00 & 54.74 & 54.49\\
 & \texttt{nca} & 15 & 0.80--2.90 & 1.30 & 53.89 & 53.50\\
 & \texttt{nca$+$ave} & 14 & 0.80--2.70 & 1.00 & 54.78 & 54.05\\
\bottomrule
\end{tabular}
\caption{Layer 1: conflict variants under contiguous slicing.}
\label{tab:grid_conflict}
\end{table}

\begin{table}[!tbp]
\centering
\footnotesize
\setlength{\tabcolsep}{2pt}
\begin{tabular}{lccccc}
\toprule
Line & \(n_\lambda\) & \(\lambda\) range & \(\lambda^{*}\) & Best & Rule\\
\midrule
\texttt{SS(ca$+$ave,energy)} & 9 & 2.00--2.75 & 2.30 & 58.21 & 57.49\\
\texttt{SS(ca$+$ave,margin)} & 9 & 2.00--2.75 & 2.30 & 57.80 & 57.62\\
\texttt{SS(ca$+$ave,mean)} & 9 & 2.00--2.75 & 2.60 & 57.98 & 57.98\\
\texttt{SS(ca$+$ave,owner)} & 9 & 2.00--2.75 & 2.20 & 57.98 & 57.76\\
\texttt{SS(ca$+$ave,variance)} & 9 & 2.00--2.75 & 2.25 & 58.08 & 57.36\\
\texttt{SS(full,energy)} & 5 & 2.30--2.70 & 2.40 & 57.97 & 57.72\\
\texttt{SS(full,margin)} & 5 & 2.30--2.70 & 2.40 & 57.76 & 57.51\\
\texttt{SS(full,mean)} & 5 & 2.30--2.70 & 2.40 & 58.08 & 57.49\\
\texttt{SS(full,owner)} & 5 & 2.30--2.70 & 2.70 & 57.84 & 57.24\\
\texttt{SS(full,variance)} & 5 & 2.30--2.70 & 2.30 & 57.99 & 57.47\\
\texttt{SS(lr,energy)} & 5 & 2.30--2.70 & 2.30 & 57.69 & 57.33\\
\texttt{SS(lr,margin)} & 5 & 2.30--2.70 & 2.50 & 57.96 & 57.53\\
\texttt{SS(lr,mean)} & 5 & 2.30--2.70 & 2.30 & 57.82 & 57.73\\
\texttt{SS(lr,owner)} & 5 & 2.30--2.70 & 2.30 & 57.92 & 57.84\\
\texttt{SS(lr,variance)} & 5 & 2.30--2.70 & 2.30 & 57.64 & 57.56\\
\texttt{SS(nca$+$ave,energy)} & 9 & 2.00--2.75 & 2.50 & 57.98 & 57.75\\
\texttt{SS(nca$+$ave,margin)} & 9 & 2.00--2.75 & 2.30 & 57.91 & 57.72\\
\texttt{SS(nca$+$ave,mean)} & 9 & 2.00--2.75 & 2.25 & 57.98 & 57.75\\
\texttt{SS(nca$+$ave,owner)} & 9 & 2.00--2.75 & 2.25 & 58.03 & 57.56\\
\texttt{SS(nca$+$ave,variance)} & 9 & 2.00--2.75 & 2.40 & 57.97 & 57.23\\
\texttt{SS(nca,energy)} & 9 & 2.00--2.75 & 2.75 & 57.85 & 57.80\\
\texttt{SS(nca,margin)} & 9 & 2.00--2.75 & 2.75 & 57.52 & 56.98\\
\texttt{SS(nca,mean)} & 9 & 2.00--2.75 & 2.30 & 57.51 & 57.37\\
\texttt{SS(nca,owner)} & 9 & 2.00--2.75 & 2.25 & 57.57 & 57.39\\
\texttt{SS(nca,variance)} & 9 & 2.00--2.75 & 2.50 & 57.63 & 57.44\\
\bottomrule
\end{tabular}
\caption{Layer 2 (T1): spread-only lines.}
\label{tab:grid_spread_t1}
\end{table}

\begin{table}[!tbp]
\centering
\footnotesize
\setlength{\tabcolsep}{2pt}
\begin{tabular}{lccccc}
\toprule
Line & \(n_\lambda\) & \(\lambda\) range & \(\lambda^{*}\) & Best & Rule\\
\midrule
\texttt{SS(ca$+$ave,energy)} & 5 & 2.60--2.90 & 2.80 & 42.83 & 42.18\\
\texttt{SS(ca$+$ave,margin)} & 5 & 2.60--2.90 & 2.70 & 42.75 & 42.44\\
\texttt{SS(ca$+$ave,mean)} & 5 & 2.60--2.90 & 2.70 & 42.77 & 42.49\\
\texttt{SS(ca$+$ave,owner)} & 5 & 2.60--2.90 & 2.75 & 42.78 & 42.51\\
\texttt{SS(ca$+$ave,variance)} & 5 & 2.60--2.90 & 2.75 & 42.80 & 41.93\\
\texttt{SS(ca,energy)} & 4 & 2.60--2.80 & 2.75 & 26.71 & 26.42\\
\texttt{SS(ca,margin)} & 4 & 2.60--2.80 & 2.60 & 26.65 & 26.65\\
\texttt{SS(ca,mean)} & 4 & 2.60--2.80 & 2.60 & 26.48 & 26.48\\
\texttt{SS(ca,owner)} & 3 & 2.60--2.75 & 2.60 & 26.60 & 26.60\\
\texttt{SS(ca,variance)} & 4 & 2.60--2.80 & 2.60 & 26.56 & 26.56\\
\texttt{SS(full,energy)} & 8 & 2.50--3.10 & 2.75 & 42.87 & 42.57\\
\texttt{SS(full,margin)} & 8 & 2.50--3.10 & 2.60 & 42.76 & 42.76\\
\texttt{SS(full,mean)} & 8 & 2.50--3.10 & 2.75 & 42.86 & 42.58\\
\texttt{SS(full,owner)} & 8 & 2.50--3.10 & 2.70 & 42.84 & 42.65\\
\texttt{SS(full,variance)} & 8 & 2.50--3.10 & 2.75 & 42.82 & 42.20\\
\texttt{SS(lr,energy)} & 8 & 2.50--3.10 & 2.80 & 42.81 & 42.58\\
\texttt{SS(lr,margin)} & 8 & 2.50--3.10 & 2.60 & 42.76 & 42.76\\
\texttt{SS(lr,mean)} & 8 & 2.50--3.10 & 2.80 & 42.99 & 42.93\\
\texttt{SS(lr,owner)} & 8 & 2.50--3.10 & 2.80 & 42.96 & 42.84\\
\texttt{SS(lr,variance)} & 8 & 2.50--3.10 & 2.75 & 42.87 & 42.72\\
\texttt{SS(nca$+$ave,energy)} & 5 & 2.60--2.90 & 2.75 & 42.66 & 42.41\\
\texttt{SS(nca$+$ave,margin)} & 5 & 2.60--2.90 & 2.75 & 42.70 & 42.41\\
\texttt{SS(nca$+$ave,mean)} & 5 & 2.60--2.90 & 2.75 & 42.69 & 42.35\\
\texttt{SS(nca$+$ave,owner)} & 5 & 2.60--2.90 & 2.80 & 42.91 & 42.37\\
\texttt{SS(nca$+$ave,variance)} & 5 & 2.60--2.90 & 2.70 & 42.80 & 42.03\\
\texttt{SS(nca,energy)} & 5 & 2.60--2.90 & 2.70 & 43.37 & 42.99\\
\texttt{SS(nca,margin)} & 5 & 2.60--2.90 & 2.75 & 42.89 & 42.77\\
\texttt{SS(nca,mean)} & 4 & 2.60--2.80 & 2.70 & 43.08 & 42.89\\
\texttt{SS(nca,owner)} & 5 & 2.60--2.90 & 2.70 & 43.01 & 42.92\\
\texttt{SS(nca,variance)} & 5 & 2.60--2.90 & 2.70 & 43.20 & 42.77\\
\bottomrule
\end{tabular}
\caption{Layer 2 (T2): spread-only lines.}
\label{tab:grid_spread_t2}
\end{table}

\begin{table}[!tbp]
\centering
\footnotesize
\setlength{\tabcolsep}{2pt}
\begin{tabular}{lccccc}
\toprule
Line & \(n_\lambda\) & \(\lambda\) range & \(\lambda^{*}\) & Best & Rule\\
\midrule
\texttt{SS(ca$+$ave,energy)} & 7 & 1.50--2.00 & 1.80 & 63.52 & 63.45\\
\texttt{SS(ca$+$ave,margin)} & 5 & 1.50--2.00 & 1.70 & 63.20 & 62.98\\
\texttt{SS(ca$+$ave,mean)} & 5 & 1.50--2.00 & 1.60 & 63.29 & 63.19\\
\texttt{SS(ca$+$ave,owner)} & 5 & 1.50--2.00 & 1.50 & 63.16 & 63.05\\
\texttt{SS(ca$+$ave,variance)} & 7 & 1.50--2.00 & 1.75 & 63.52 & 63.27\\
\texttt{SS(full,energy)} & 7 & 1.50--2.00 & 1.75 & 63.31 & 63.30\\
\texttt{SS(full,margin)} & 7 & 1.50--2.00 & 1.60 & 63.42 & 63.11\\
\texttt{SS(full,mean)} & 7 & 1.50--2.00 & 2.00 & 63.19 & 63.19\\
\texttt{SS(full,owner)} & 7 & 1.50--2.00 & 1.75 & 63.26 & 63.11\\
\texttt{SS(full,variance)} & 7 & 1.50--2.00 & 2.00 & 63.42 & 63.42\\
\texttt{SS(lr,energy)} & 7 & 1.50--2.00 & 1.80 & 63.38 & 63.10\\
\texttt{SS(lr,margin)} & 7 & 1.50--2.00 & 2.00 & 63.33 & 63.33\\
\texttt{SS(lr,mean)} & 7 & 1.50--2.00 & 1.90 & 63.23 & 63.23\\
\texttt{SS(lr,owner)} & 7 & 1.50--2.00 & 2.00 & 63.34 & 63.34\\
\texttt{SS(lr,variance)} & 7 & 1.50--2.00 & 1.90 & 63.41 & 63.02\\
\texttt{SS(nca$+$ave,energy)} & 7 & 1.50--2.00 & 1.80 & 63.57 & 63.34\\
\texttt{SS(nca$+$ave,margin)} & 5 & 1.50--2.00 & 1.75 & 63.31 & 63.24\\
\texttt{SS(nca$+$ave,mean)} & 5 & 1.50--2.00 & 2.00 & 63.57 & 63.57\\
\texttt{SS(nca$+$ave,owner)} & 5 & 1.50--2.00 & 2.00 & 63.39 & 63.39\\
\texttt{SS(nca$+$ave,variance)} & 7 & 1.50--2.00 & 1.75 & 63.73 & 63.17\\
\texttt{SS(nca,energy)} & 7 & 1.50--2.00 & 1.60 & 63.72 & 63.12\\
\texttt{SS(nca,margin)} & 5 & 1.50--2.00 & 2.00 & 63.34 & 63.34\\
\texttt{SS(nca,mean)} & 5 & 1.50--2.00 & 1.70 & 63.17 & 62.95\\
\texttt{SS(nca,owner)} & 5 & 1.50--2.00 & 1.70 & 63.21 & 63.14\\
\texttt{SS(nca,variance)} & 7 & 1.50--2.00 & 1.50 & 63.66 & 63.27\\
\bottomrule
\end{tabular}
\caption{Layer 2 (T3): spread-only lines.}
\label{tab:grid_spread_t3}
\end{table}

\begin{table}[!tbp]
\centering
\footnotesize
\setlength{\tabcolsep}{2pt}
\begin{tabular}{lccccc}
\toprule
Line & \(n_\lambda\) & \(\lambda\) range & \(\lambda^{*}\) & Best & Rule\\
\midrule
\texttt{SS(ca$+$ave,energy)} & 5 & 1.00--1.40 & 1.00 & 53.72 & 53.68\\
\texttt{SS(ca$+$ave,margin)} & 6 & 1.00--1.40 & 1.10 & 53.28 & 53.28\\
\texttt{SS(ca$+$ave,mean)} & 6 & 1.00--1.40 & 1.00 & 53.72 & 53.38\\
\texttt{SS(ca$+$ave,owner)} & 5 & 1.10--1.40 & 1.10 & 53.67 & 53.67\\
\texttt{SS(ca$+$ave,variance)} & 6 & 1.00--1.40 & 1.00 & 53.51 & 53.49\\
\texttt{SS(full,energy)} & 5 & 1.00--1.30 & 1.00 & 53.29 & 53.17\\
\texttt{SS(full,margin)} & 5 & 1.00--1.30 & 1.10 & 53.59 & 53.59\\
\texttt{SS(full,mean)} & 5 & 1.00--1.30 & 1.10 & 53.57 & 53.57\\
\texttt{SS(full,owner)} & 5 & 1.00--1.30 & 1.00 & 53.84 & 53.46\\
\texttt{SS(full,variance)} & 5 & 1.00--1.30 & 1.00 & 53.57 & 53.23\\
\texttt{SS(lr,energy)} & 5 & 1.00--1.30 & 1.10 & 53.58 & 53.58\\
\texttt{SS(lr,margin)} & 5 & 1.00--1.30 & 1.25 & 53.98 & 53.81\\
\texttt{SS(lr,mean)} & 5 & 1.00--1.30 & 1.00 & 53.69 & 53.58\\
\texttt{SS(lr,owner)} & 5 & 1.00--1.30 & 1.00 & 54.16 & 54.15\\
\texttt{SS(lr,variance)} & 4 & 1.00--1.25 & 1.00 & 53.99 & 53.46\\
\texttt{SS(nca$+$ave,energy)} & 6 & 1.00--1.40 & 1.25 & 53.92 & 53.17\\
\texttt{SS(nca$+$ave,margin)} & 6 & 1.00--1.40 & 1.10 & 53.77 & 53.77\\
\texttt{SS(nca$+$ave,mean)} & 6 & 1.00--1.40 & 1.00 & 53.65 & 53.48\\
\texttt{SS(nca$+$ave,owner)} & 6 & 1.00--1.40 & 1.00 & 53.58 & 53.53\\
\texttt{SS(nca$+$ave,variance)} & 6 & 1.00--1.40 & 1.00 & 53.30 & 53.16\\
\texttt{SS(nca,energy)} & 6 & 1.00--1.40 & 1.10 & 53.70 & 53.70\\
\texttt{SS(nca,margin)} & 6 & 1.00--1.40 & 1.10 & 52.66 & 52.66\\
\texttt{SS(nca,mean)} & 6 & 1.00--1.40 & 1.10 & 52.83 & 52.83\\
\texttt{SS(nca,owner)} & 5 & 1.00--1.40 & 1.25 & 52.87 & 52.52\\
\texttt{SS(nca,variance)} & 6 & 1.00--1.40 & 1.25 & 52.74 & 52.66\\
\bottomrule
\end{tabular}
\caption{Layer 2 (T4): spread-only lines.}
\label{tab:grid_spread_t4}
\end{table}

\begin{table}[!tbp]
\centering
\footnotesize
\setlength{\tabcolsep}{3pt}
\begin{tabular}{llccccc}
\toprule
Suite & Line & \(n_\lambda\) & \(\lambda\) range & \(\lambda^{*}\) & Best & Rule\\
\midrule
T2 & \texttt{ca$+$ave.ta} & 12 & 2.10--3.00 & 2.55 & 42.77 & 42.61\\
 & \texttt{ca$+$ave.ties} & 12 & 2.10--3.00 & 2.55 & 42.96 & 42.96\\
 & \texttt{ca$+$ave.tsvm} & 12 & 2.10--3.00 & 2.65 & 42.83 & 42.65\\
 & \texttt{full.ta} & 8 & 2.20--2.90 & 2.60 & 42.72 & 42.72\\
 & \texttt{full.ties} & 8 & 2.20--2.90 & 2.60 & 43.03 & 43.03\\
 & \texttt{full.tsvm} & 8 & 2.20--2.90 & 2.60 & 42.79 & 42.79\\
 & \texttt{lr.ta} & 8 & 2.20--2.90 & 2.60 & 42.82 & 42.82\\
 & \texttt{lr.ties} & 8 & 2.20--2.90 & 2.60 & 43.11 & 43.11\\
 & \texttt{lr.tsvm} & 8 & 2.20--2.90 & 2.80 & 43.06 & 42.99\\
 & \texttt{nca$+$ave.ta} & 12 & 2.10--3.00 & 2.65 & 42.76 & 42.50\\
 & \texttt{nca$+$ave.ties} & 12 & 2.10--3.00 & 2.55 & 43.18 & 42.69\\
 & \texttt{nca$+$ave.tsvm} & 12 & 2.10--3.00 & 2.55 & 42.98 & 42.90\\
 & \texttt{nca.ta} & 12 & 2.10--3.00 & 2.60 & 43.00 & 43.00\\
 & \texttt{nca.ties} & 12 & 2.10--3.00 & 2.60 & 43.32 & 43.32\\
 & \texttt{nca.tsvm} & 12 & 2.10--3.00 & 2.80 & 43.06 & 42.94\\
T3 & \texttt{ca$+$ave.ta} & 7 & 1.50--2.00 & 1.70 & 63.40 & 63.40\\
 & \texttt{ca$+$ave.ties} & 7 & 1.50--2.00 & 1.50 & 62.71 & 62.29\\
 & \texttt{ca$+$ave.tsvm} & 7 & 1.50--2.00 & 1.90 & 63.10 & 62.55\\
 & \texttt{full.ta} & 7 & 1.50--2.00 & 1.80 & 63.61 & 63.46\\
 & \texttt{full.ties} & 7 & 1.50--2.00 & 1.75 & 62.90 & 62.25\\
 & \texttt{full.tsvm} & 7 & 1.50--2.00 & 1.75 & 63.04 & 62.94\\
 & \texttt{lr.ta} & 7 & 1.50--2.00 & 2.00 & 63.64 & 63.64\\
 & \texttt{lr.ties} & 7 & 1.50--2.00 & 1.50 & 63.10 & 62.21\\
 & \texttt{lr.tsvm} & 7 & 1.50--2.00 & 1.50 & 63.19 & 63.10\\
 & \texttt{nca$+$ave.ta} & 7 & 1.50--2.00 & 1.70 & 63.60 & 63.28\\
 & \texttt{nca$+$ave.ties} & 7 & 1.50--2.00 & 1.50 & 63.13 & 62.21\\
 & \texttt{nca$+$ave.tsvm} & 7 & 1.50--2.00 & 1.50 & 62.94 & 62.70\\
 & \texttt{nca.ta} & 7 & 1.50--2.00 & 1.75 & 63.75 & 63.46\\
 & \texttt{nca.ties} & 7 & 1.50--2.00 & 1.60 & 63.05 & 62.57\\
 & \texttt{nca.tsvm} & 7 & 1.50--2.00 & 1.90 & 63.20 & 62.56\\
T4 & \texttt{ca$+$ave.ta} & 6 & 1.00--1.40 & 1.00 & 54.46 & 53.56\\
 & \texttt{ca$+$ave.ties} & 6 & 1.00--1.40 & 1.00 & 54.67 & 54.24\\
 & \texttt{ca$+$ave.tsvm} & 6 & 1.00--1.40 & 1.30 & 52.40 & 51.67\\
 & \texttt{full.ta} & 16 & 1.00--2.90 & 1.20 & 54.48 & 54.36\\
 & \texttt{full.ties} & 16 & 1.00--2.90 & 1.20 & 54.67 & 54.29\\
 & \texttt{full.tsvm} & 16 & 1.00--2.90 & 1.20 & 53.42 & 52.76\\
 & \texttt{lr.ta} & 16 & 1.00--2.90 & 1.10 & 54.48 & 54.48\\
 & \texttt{lr.ties} & 16 & 1.00--2.90 & 1.25 & 54.84 & 54.56\\
 & \texttt{lr.tsvm} & 13 & 1.00--2.80 & 1.25 & 53.30 & 53.07\\
 & \texttt{nca$+$ave.ta} & 6 & 1.00--1.40 & 1.00 & 54.43 & 53.68\\
 & \texttt{nca$+$ave.ties} & 6 & 1.00--1.40 & 1.20 & 54.72 & 54.37\\
 & \texttt{nca$+$ave.tsvm} & 6 & 1.00--1.40 & 1.30 & 53.08 & 51.97\\
 & \texttt{nca.ta} & 8 & 1.00--1.60 & 1.10 & 53.98 & 53.98\\
 & \texttt{nca.ties} & 7 & 1.00--1.50 & 1.30 & 53.86 & 53.57\\
 & \texttt{nca.tsvm} & 8 & 1.00--1.60 & 1.30 & 53.21 & 51.16\\
\bottomrule
\end{tabular}
\caption{Layer 3: residual-pathway lines without spread slicing.}
\label{tab:grid_rp}
\end{table}

\begin{table}[!tbp]
\centering
\scriptsize
\setlength{\tabcolsep}{2pt}
\begin{tabular}{lccccc}
\toprule
Line & \(n_\lambda\) & \(\lambda\) range & \(\lambda^{*}\) & Best & Rule\\
\midrule
\texttt{SS(ca$+$ave,energy).ties} & 4 & 2.60--2.80 & 2.60 & 43.34 & 43.34\\
\texttt{SS(ca$+$ave,margin).ties} & 4 & 2.60--2.80 & 2.60 & 43.22 & 43.22\\
\texttt{SS(ca$+$ave,mean).ties} & 4 & 2.60--2.80 & 2.60 & 42.72 & 42.72\\
\texttt{SS(ca$+$ave,owner).ties} & 4 & 2.60--2.80 & 2.60 & 42.77 & 42.77\\
\texttt{SS(ca$+$ave,variance).ties} & 4 & 2.60--2.80 & 2.60 & 43.48 & 43.48\\
\texttt{SS(full,energy).ties} & 4 & 2.60--2.80 & 2.60 & 43.07 & 43.07\\
\texttt{SS(full,margin).ties} & 4 & 2.60--2.80 & 2.70 & 42.73 & 42.72\\
\texttt{SS(full,mean).ties} & 4 & 2.60--2.80 & 2.60 & 42.98 & 42.98\\
\texttt{SS(full,owner).ties} & 4 & 2.60--2.80 & 2.60 & 42.88 & 42.88\\
\texttt{SS(full,variance).ties} & 4 & 2.60--2.80 & 2.70 & 43.04 & 42.85\\
\texttt{SS(lr,energy).ties} & 4 & 2.60--2.80 & 2.60 & 43.09 & 43.09\\
\texttt{SS(lr,margin).ties} & 4 & 2.60--2.80 & 2.70 & 43.05 & 43.01\\
\texttt{SS(lr,mean).ties} & 4 & 2.60--2.80 & 2.60 & 43.30 & 43.30\\
\texttt{SS(lr,owner).ties} & 4 & 2.60--2.80 & 2.60 & 43.07 & 43.07\\
\texttt{SS(lr,variance).ties} & 4 & 2.60--2.80 & 2.60 & 42.99 & 42.99\\
\texttt{SS(nca$+$ave,energy).ties} & 4 & 2.60--2.80 & 2.60 & 42.95 & 42.95\\
\texttt{SS(nca$+$ave,margin).ties} & 4 & 2.60--2.80 & 2.70 & 42.93 & 42.89\\
\texttt{SS(nca$+$ave,mean).ties} & 4 & 2.60--2.80 & 2.60 & 42.96 & 42.96\\
\texttt{SS(nca$+$ave,owner).ties} & 4 & 2.60--2.80 & 2.75 & 43.14 & 42.75\\
\texttt{SS(nca$+$ave,variance).ties} & 4 & 2.60--2.80 & 2.60 & 43.30 & 43.30\\
\texttt{SS(nca,energy).ties} & 7 & 2.20--2.80 & 2.80 & 43.15 & 42.84\\
\texttt{SS(nca,margin).ties} & 4 & 2.60--2.80 & 2.60 & 43.39 & 43.39\\
\texttt{SS(nca,mean).ties} & 4 & 2.60--2.80 & 2.60 & 43.28 & 43.28\\
\texttt{SS(nca,owner).ties} & 4 & 2.60--2.80 & 2.60 & 43.28 & 43.28\\
\texttt{SS(nca,variance).ties} & 8 & 2.20--2.80 & 2.60 & 43.46 & 43.46\\
\bottomrule
\end{tabular}
\caption{Layer 4 (T2): combined spread$+$residual lines.}
\label{tab:grid_ssrp_t2}
\end{table}

\begin{table}[!tbp]
\centering
\scriptsize
\setlength{\tabcolsep}{2pt}
\begin{tabular}{lccccc}
\toprule
Line & \(n_\lambda\) & \(\lambda\) range & \(\lambda^{*}\) & Best & Rule\\
\midrule
\texttt{SS(ca$+$ave,energy).ta} & 6 & 1.50--2.00 & 1.70 & 63.42 & 63.13\\
\texttt{SS(ca$+$ave,margin).ta} & 6 & 1.50--2.00 & 1.70 & 63.66 & 63.43\\
\texttt{SS(ca$+$ave,mean).ta} & 6 & 1.50--2.00 & 1.70 & 63.64 & 63.12\\
\texttt{SS(ca$+$ave,owner).ta} & 6 & 1.50--2.00 & 1.70 & 63.53 & 63.16\\
\texttt{SS(ca$+$ave,variance).ta} & 6 & 1.50--2.00 & 1.70 & 63.59 & 63.29\\
\texttt{SS(nca$+$ave,energy).ta} & 6 & 1.50--2.00 & 1.80 & 63.56 & 63.30\\
\texttt{SS(nca$+$ave,margin).ta} & 6 & 1.50--2.00 & 1.70 & 63.49 & 63.11\\
\texttt{SS(nca$+$ave,mean).ta} & 6 & 1.50--2.00 & 1.75 & 63.59 & 63.45\\
\texttt{SS(nca$+$ave,owner).ta} & 6 & 1.50--2.00 & 1.80 & 63.43 & 63.39\\
\texttt{SS(nca$+$ave,variance).ta} & 6 & 1.50--2.00 & 1.70 & 63.56 & 63.17\\
\texttt{SS(nca,energy).ta} & 6 & 1.50--2.00 & 1.70 & 63.59 & 63.18\\
\texttt{SS(nca,margin).ta} & 6 & 1.50--2.00 & 1.70 & 63.54 & 63.33\\
\texttt{SS(nca,mean).ta} & 6 & 1.50--2.00 & 1.60 & 63.52 & 63.04\\
\texttt{SS(nca,owner).ta} & 6 & 1.50--2.00 & 1.70 & 63.47 & 63.26\\
\texttt{SS(nca,variance).ta} & 6 & 1.50--2.00 & 1.70 & 63.58 & 63.52\\
\bottomrule
\end{tabular}
\caption{Layer 4 (T3): combined spread$+$residual lines.}
\label{tab:grid_ssrp_t3}
\end{table}

\begin{table}[!tbp]
\centering
\footnotesize
\setlength{\tabcolsep}{2pt}
\begin{tabular}{lccccc}
\toprule
Line & \(n_\lambda\) & \(\lambda\) range & \(\lambda^{*}\) & Best & Rule\\
\midrule
\texttt{SS(ca$+$ave,energy).ta} & 6 & 1.00--1.40 & 1.00 & 54.49 & 54.11\\
\texttt{SS(ca$+$ave,margin).ta} & 6 & 1.00--1.40 & 1.25 & 54.51 & 53.96\\
\texttt{SS(ca$+$ave,mean).ta} & 6 & 1.00--1.40 & 1.10 & 53.85 & 53.85\\
\texttt{SS(ca$+$ave,owner).ta} & 4 & 1.00--1.40 & 1.00 & 53.87 & 53.81\\
\texttt{SS(ca$+$ave,variance).ta} & 6 & 1.00--1.40 & 1.00 & 54.46 & 54.36\\
\texttt{SS(full,energy).ta} & 6 & 1.00--1.40 & 1.00 & 54.41 & 53.90\\
\texttt{SS(full,margin).ta} & 6 & 1.00--1.40 & 1.10 & 54.67 & 54.67\\
\texttt{SS(full,mean).ta} & 6 & 1.00--1.40 & 1.00 & 53.95 & 53.47\\
\texttt{SS(full,owner).ta} & 6 & 1.00--1.40 & 1.00 & 53.87 & 53.77\\
\texttt{SS(full,variance).ta} & 6 & 1.00--1.40 & 1.20 & 54.70 & 54.00\\
\texttt{SS(lr,energy).ta} & 6 & 1.00--1.40 & 1.20 & 54.56 & 54.51\\
\texttt{SS(lr,margin).ta} & 6 & 1.00--1.40 & 1.00 & 54.86 & 54.50\\
\texttt{SS(lr,mean).ta} & 6 & 1.00--1.40 & 1.25 & 54.99 & 54.62\\
\texttt{SS(lr,owner).ta} & 6 & 1.00--1.40 & 1.10 & 53.97 & 53.97\\
\texttt{SS(lr,variance).ta} & 6 & 1.00--1.40 & 1.00 & 54.73 & 54.71\\
\texttt{SS(nca$+$ave,energy).ta} & 7 & 1.00--1.60 & 1.20 & 54.44 & 54.41\\
\texttt{SS(nca$+$ave,margin).ta} & 6 & 1.00--1.40 & 1.00 & 54.84 & 53.58\\
\texttt{SS(nca$+$ave,mean).ta} & 6 & 1.00--1.40 & 1.00 & 54.33 & 53.36\\
\texttt{SS(nca$+$ave,owner).ta} & 5 & 1.00--1.30 & 1.00 & 54.08 & 53.71\\
\texttt{SS(nca$+$ave,variance).ta} & 6 & 1.00--1.40 & 1.30 & 53.94 & 53.89\\
\texttt{SS(nca,energy).ta} & 7 & 1.00--1.60 & 1.10 & 53.83 & 53.83\\
\texttt{SS(nca,margin).ta} & 7 & 1.00--1.50 & 1.20 & 53.62 & 53.40\\
\texttt{SS(nca,mean).ta} & 6 & 1.00--1.40 & 1.10 & 53.30 & 53.30\\
\texttt{SS(nca,owner).ta} & 6 & 1.00--1.40 & 1.10 & 53.22 & 53.22\\
\texttt{SS(nca,variance).ta} & 6 & 1.00--1.40 & 1.10 & 53.93 & 53.93\\
\bottomrule
\end{tabular}
\caption{Layer 4 (T4): combined spread$+$residual lines.}
\label{tab:grid_ssrp_t4}
\end{table}

\subsection{Best Results at the \(\kappa\)-Rule Point and after Sweeping}

Table~\ref{tab:best_summary} separates the result selected by the
\(\kappa\)-rule from the best value found by sweeping \(\lambda\).

\begin{table*}[!tp]
\centering
\scriptsize
\setlength{\tabcolsep}{4pt}
\begin{tabular}{lllllllll}
\toprule
Suite
& Rule-selected line
& Rule \(\lambda\)
& Rule Avg
& Swept-best line
& Best \(\lambda\)
& Best Avg
& Best OM
& Best WS\\
\midrule
T1
& \texttt{SS(ca$+$ave,mean)}
& 2.60
& 57.98
& \texttt{SS(ca$+$ave,energy)}
& 2.30
& 58.21
& 57.13
& 56.23\\
T2
& \texttt{SS(ca$+$ave,variance).ties}
& 2.60
& 43.48
& \texttt{SS(ca$+$ave,variance).ties}
& 2.60
& 43.48
& 42.13
& 43.45\\
T3
& \texttt{lr.ta}
& 2.00
& 64.38
& \texttt{lr.ta}
& 2.00
& 64.38
& 63.56
& 63.70\\
T4
& \texttt{SS(lr,variance).ta}
& 1.10
& 54.71
& \texttt{SS(lr,mean).ta}
& 1.25
& 54.99
& 54.74
& 54.57\\
\bottomrule
\end{tabular}
\caption{Best in-domain averages at the \(\kappa\)-rule point and at
the swept optimum, with the internal line
(Table~\ref{tab:config_names}) realizing each value. The
rule-selected column reports the best configuration evaluated
directly at the \(\kappa\)-rule point. It may correspond to a
different internal line than the swept best within the same
component family (T1 and T4 here), and may differ from the
main-table line, which is selected by its swept optimum
(Table~\ref{tab:main_mapping}). Best OM is the strongest OrthoMerge
variant and Best WS the strongest weight-space baseline for each
suite, from the main tables. The rule-selected CORAM exceeds the
strongest OrthoMerge variant on T1--T3; against the strongest
weight-space baseline it leads on T1 and T3 and is comparable on T2
(43.48 vs.\ 43.45, within evaluation noise; Section~H). On T4 it
trails OrthoMerge by \(0.03\) at the rule point and exceeds it
after the sweep.}
\label{tab:best_summary}
\end{table*}

\subsection{Distribution of the Swept Optima}

On T4, the swept optima of the retained lines are distributed as
follows:

\begin{table}[!tbp]
\centering
\small
\begin{tabular}{lccccc}
\toprule
\(\lambda^{*}\)
& 1.00 & 1.10 & 1.20 & 1.25 & 1.30\\
\midrule
Number of lines
& 16 & 10 & 9 & 4 & 6\\
\bottomrule
\end{tabular}
\caption{Distribution of the swept-optimal coefficient across the
retained T4 lines.}
\label{tab:t4_lambda_distribution}
\end{table}

The median is \(\lambda^{*}=1.10\), corresponding to
\[
\kappa_{\mathrm{opt}}
=
\frac{1.10}{\sqrt{5}}
\approx0.49.
\]
The swept values span approximately
\(
\kappa\in[0.45,0.58].
\)
We therefore use the rounded shared constant \(\kappa=0.5\) for the
high-dispersion case.

For the low-dispersion case, the main-table configurations give
approximately
\[
\kappa_{\mathrm{opt}}\in[1.03,1.12]
\quad\text{on T1},
\]
\[
\kappa_{\mathrm{opt}}\in[1.12,1.21]
\quad\text{on T2},
\]
and
\[
\kappa_{\mathrm{opt}}\in[1.01,1.16]
\quad\text{on T3}.
\]
The single swept Gemma-2-2B configuration gives
\(
\kappa_{\mathrm{opt}}\approx1.07.
\)
The shared value \(\kappa=1.15\) lies within or near these ranges and
maps to \(\lambda=2.60\) for \(N=5\) and \(\lambda=2.00\) for \(N=3\).
The swept curves are flat around their optima; using
\(\kappa=1.10\) does not change the main comparisons.

\subsection{Threshold Sensitivity}

The rule assigns \(\kappa\) by thresholding \(D\). The observed values
leave a wide unpopulated gap: the low-dispersion bases have
\(D\in[1.3,3.5]\) and the single high-dispersion base has
\(D\approx16\). Every threshold \(D_0\) in the open interval
\((3.5,16)\) therefore produces the same assignment, the same selected
coefficients, and the same rule-to-optimum gaps
(Table~\ref{tab:threshold_sens}).

\begin{table}[!tbp]
\centering
\small
\setlength{\tabcolsep}{5pt}
\begin{tabular}{lccc}
\toprule
\(D_0\) & Assignments & Per-suite gaps (T1--T4) & Max gap\\
\midrule
4  & unchanged & 0.23 / 0.00 / 0.00 / 0.28 & 0.28\\
6  & unchanged & 0.23 / 0.00 / 0.00 / 0.28 & 0.28\\
8  & unchanged & 0.23 / 0.00 / 0.00 / 0.28 & 0.28\\
10 & unchanged & 0.23 / 0.00 / 0.00 / 0.28 & 0.28\\
12 & unchanged & 0.23 / 0.00 / 0.00 / 0.28 & 0.28\\
\bottomrule
\end{tabular}
\caption{Threshold sensitivity. Gaps are the best-configuration
rule-to-optimum differences of Table~\ref{tab:best_summary}. Any
\(D_0\in(3.5,16)\) yields identical behavior. We fix no unique
threshold; intermediate values of \(D\) are not represented in the
current expert suites (see Limitations).}
\label{tab:threshold_sens}
\end{table}

\subsection{Leave-One-Base-Out Validation}

For each low-dispersion base we re-estimate \(\kappa\) using only the
remaining low-dispersion bases and test the resulting coefficient on
the held-out base. The representative \(\kappa_{\mathrm{opt}}\) of a
base is the midpoint of its per-configuration range reported above.
The held-out estimate \(\hat{\kappa}\) is the mean over the other
three bases, including Gemma-2-2B. The predicted coefficient
\(\lambda_{\mathrm{pred}}=\hat{\kappa}\sqrt{N}\) is rounded to the
nearest evaluated grid point.

\begin{table}[!tbp]
\centering
\footnotesize
\setlength{\tabcolsep}{3pt}
\begin{tabular}{lccccc}
\toprule
Held-out & \(\hat{\kappa}\) & \(\lambda_{\mathrm{pred}}\)
& Score & Best & Gap\\
\midrule
T1 (\texttt{full}) & 1.107 & 2.48\,\(\to\)\,2.50 & 57.94 & 57.94 & 0.00\\
T2 (\texttt{full}) & 1.077 & 2.41\,\(\to\)\,2.50 & 42.47 & 42.76 & 0.29\\
T3 (\texttt{lr})   & 1.103 & 1.91\,\(\to\)\,1.90 & 63.07 & 63.43 & 0.36\\
\bottomrule
\end{tabular}
\caption{Leave-one-base-out validation on the low-dispersion bases at
\(h=8\). T3 entries are five-task averages. Predicted coefficients
are rounded to the nearest point of the original evaluation grids;
for T2 this is \(\lambda{=}2.50\) (a later direct \(\lambda{=}2.40\)
evaluation on newer hardware is not comparable to the original
sweep and is not used). Only one high-dispersion base exists, so no
leave-one-out test is possible for \(\kappa=0.5\), which remains a
single-base calibration (see Limitations).}
\label{tab:loo}
\end{table}

\subsection{Alternative Coefficient Choices}

Before adopting \(\lambda=\kappa\sqrt{N}\), we tested several
closed-form alternatives over grids within \(\lambda\in[1,4]\).
Table~\ref{tab:alternative_lambda} summarizes their predictions and
observed limitations.

\begin{table*}[!tp]
\centering
\small
\setlength{\tabcolsep}{5pt}
\begin{tabular}{llll}
\toprule
Family
& Statistic
& Prediction
& Observation\\
\midrule
Weight norm
& \(\|\Delta\|\)-matching
& \(\lambda\approx1\)
& Below the swept optima\\
Energy ratio
& \(E_{\mathrm{task}}/E_{\mathrm{geo}}\)
& \(2.08/2.07/1.75\)
& Correct ordering; under-scaled\\
Spectral
& \(\sqrt{c}\)
& \(1.08/2.73/2.55\)
& Matches T2 only\\
Gram
& Cross-term statistic
& \(1.92/2.13/1.72\)
& Incorrect direction\\
Fisher energy
& Function-space energy
& \(1.87/1.81\)
& Under-scaled\\
KL path
& Calibration divergence
& \(\lambda\le1.5\)
& Below useful range\\
Linear in \(N\)
& \((N+1)/2\)
& \(3.0/3.0/2.0\)
& Overshoots for \(N=5\)\\
Geodesic
& Curvature correction
& \(2.09/2.03/1.72\)
& Below linear estimate\\
Cancellation
& \(\sqrt{N}\)
& \(2.24/2.24/1.73\)
& Best common scale\\
\bottomrule
\end{tabular}
\caption{Alternative closed-form choices considered for the
amplification scale.}
\label{tab:alternative_lambda}
\end{table*}

\section{E. Component Ablations}
\label{app:ablations}

This section separates the effects of conflict handling, spread
slicing, and the residual pathway. Table~\ref{tab:component_summary}
first reports the best value within each component branch at the
\(\kappa\)-rule point. The matched comparisons below keep the slice
height, geometric branch, spread priority, residual merger, and
amplification coefficient fixed, except for the component being
tested.

\begin{table}[!tbp]
\centering
\small
\setlength{\tabcolsep}{4pt}
\begin{tabular}{lcccc}
\toprule
Configuration & T1 & T2 & T3 & T4\\
\midrule
CORAM
& 57.72 & 42.75 & 63.95 & 54.49\\
+RP
& n/a & 42.31 & 64.38 & 54.56\\
+SS
& 57.84 & 42.93 & 63.96 & 54.15\\
+SS+RP
& n/a & 43.30 & 64.27 & 54.71\\
\midrule
CORAM-C
& 57.60 & 42.96 & 64.02 & 54.32\\
+RP
& n/a & 43.32 & 64.20 & 54.37\\
+SS
& 57.98 & 42.99 & 64.14 & 53.77\\
+SS+RP
& n/a & 43.48 & 64.33 & 54.41\\
\bottomrule
\end{tabular}
\caption{Best in-domain average within each component branch at the
\(\kappa\)-rule point. These are branch-level summaries; the matched
ablations are reported below. Branch-level bests search over all
spread priorities and conflict variants evaluated at the rule point,
including combinations not retained in the per-line tables of
Section~D; the main-table rows instead report, for each branch, the
line with the best swept optimum, evaluated at the rule coefficient.
The two conventions can differ: on T1 the CORAM-C$+$SS branch best at
the rule point is SS(ca$+$ave,mean) at 57.98, while the main-table
CORAM-C$+$SS row reports SS(ca$+$ave,energy) at 57.49, whose swept
optimum (58.21) is the strongest in the branch.}
\label{tab:component_summary}
\end{table}

\subsection{Conflict-Aware Merging}

Table~\ref{tab:conflict_ablation} compares plain and conflict-aware
merging under matched settings.

\begin{table}[!tbp]
\centering
\footnotesize
\setlength{\tabcolsep}{4pt}
\begin{tabular}{llccc}
\toprule
Suite
& Matched setting
& Plain
& Conflict-aware
& \(\Delta\)\\
\midrule
T1 & \(h{=}8\), base, \(\lambda{=}2.60\)
& 57.72
& 57.10
& \(-0.62\)\\
T2 & \(h{=}8\), base, \(\lambda{=}2.60\)
& 42.75
& 42.33
& \(-0.42\)\\
T3 & \(h{=}8\), base, \(\lambda{=}2.00\) 
& 63.85
& 64.02
& \(+0.17\)\\
T4 & \(h{=}8\), base, \(\lambda{=}1.10\)
& 54.33
& 54.05
& \(-0.28\)\\
\bottomrule
\end{tabular}
\caption{Matched conflict-aware ablation: plain \texttt{full} vs.\
conflict-aware \texttt{nca+ave}, \(\kappa\)-rule \(\lambda\), no SS or
RP. T3 averages include CharXiv.}
\label{tab:conflict_ablation}
\end{table}

\subsection{Spread Slicing}

Table~\ref{tab:spread_cross_suite} compares the strongest contiguous
and spread-only lines within the same branch.

\begin{table}[!tbp]
\centering
\small
\begin{tabular}{lccc}
\toprule
Suite or branch
& Contiguous
& Spread-only
& \(\Delta\)\\
\midrule
T1
& 57.87
& 58.21
& \(+0.34\)\\
T2
& 42.96
& 43.37
& \(+0.41\)\\
T3
& 64.04
& 64.14
& \(+0.10\)\\
T4, lr
& 54.74
& 54.16
& \(-0.58\)\\
T4, full
& 54.56
& 53.84
& \(-0.72\)\\
T4, nca+ave
& 54.78
& 53.92
& \(-0.86\)\\
\bottomrule
\end{tabular}
\caption{Best contiguous and spread-only lines within the same branch
at \(h=8\), each at its own swept optimum. T3 averages include CharXiv.
The matched fixed-coefficient comparison is in
Table~\ref{tab:spread_matched}.}
\label{tab:spread_cross_suite}
\end{table}

The final matched ablation uses the same \(\lambda\), \(h\),
geometric branch, and conflict setting for the contiguous and spread
variants:

\begin{table*}[!tp]
\centering
\small
\begin{tabular}{llllllll}
\toprule
Suite
& Branch
& \(h\)
& \(\lambda\)
& Priority
& Contiguous
& Spread
& \(\Delta\)\\
\midrule
T1
& \texttt{full}
& 8
& 2.60
& energy
& 57.72
& 57.72
& \(0.00\)\\
T2
& \texttt{full}
& 8
& 2.60
& energy
& 42.75
& 42.57
& \(-0.18\)\\
T3
& \texttt{full}
& 8
& 2.00
& energy
& 63.85
& 63.87
& \(+0.02\)\\
T4
& \texttt{full}
& 8
& 1.10
& energy
& 54.33
& 53.17
& \(-1.16\)\\
\bottomrule
\end{tabular}
\caption{Matched spread-slicing ablation: \texttt{full} branch, energy
priority, \(\kappa\)-rule \(\lambda\). T3 averages include CharXiv.
The effect is negligible on T1--T3 and negative on T4.}
\label{tab:spread_matched}
\end{table*}

\subsection{Residual Pathway}

Table~\ref{tab:rp_ablation} compares each geometric merge with and
without the residual pathway under matched settings.

\begin{table*}[!tp]
\centering
\small
\begin{tabular}{llllllll}
\toprule
Suite
& Geometry
& \(h\)
& \(\lambda\)
& Residual merger
& Without RP
& With RP
& \(\Delta\)\\
\midrule
T2
& \texttt{full}
& 8
& 2.60
& TIES
& 42.75
& 43.03
& \(+0.28\)\\
T3
& \texttt{full}
& 8
& 2.00
& Task arithmetic
& 63.85
& 64.22
& \(+0.37\)\\
T4
& \texttt{full}
& 8
& 1.10
& TIES
& 54.33
& 54.29
& \(-0.04\)\\
\bottomrule
\end{tabular}
\caption{Matched residual-pathway ablation: \texttt{full} branch,
\(\kappa\)-rule \(\lambda\). T3 averages include CharXiv.}
\label{tab:rp_ablation}
\end{table*}

\subsection{Interaction between Spread Slicing and the Residual Pathway}

The four-way comparison in Table~\ref{tab:ss_rp_interaction} separates
the individual effects of spread slicing and the residual pathway.

\begin{table}[!tbp]
\centering
\footnotesize
\setlength{\tabcolsep}{4pt}
\begin{tabular}{llllll}
\toprule
Suite
& Base branch
& Neither
& SS only
& RP only
& SS+RP\\
\midrule
T2
& CORAM
& 42.75
& 42.93
& 42.31
& 43.30\\
T2
& CORAM-C
& 42.96
& 42.99
& 43.32
& 43.48\\
T3
& CORAM
& 63.95
& 63.96
& 64.38
& 64.27\\
T3
& CORAM-C
& 64.02
& 64.14
& 64.20
& 64.33\\
T4
& CORAM
& 54.49
& 54.15
& 54.56
& 54.71\\
T4
& CORAM-C
& 54.32
& 53.77
& 54.37
& 54.41\\
\bottomrule
\end{tabular}
\caption{Branch-level interaction at the \(\kappa\)-rule point, with
the best priority and merger per suite and branch. Matched comparisons:
Tables~\ref{tab:spread_matched} and~\ref{tab:rp_ablation}.}
\label{tab:ss_rp_interaction}
\end{table}

\subsection{Additional Analysis of Spread Slicing on T4}

\paragraph{Task-level effect.}
For a matched owner-balance/full comparison at \(\lambda=1.20\), the
changes are
\[
-0.93 \text{ on Instruction},\quad
+0.07 \text{ on Math},\quad
+0.49 \text{ on Coding},
\]
\[
+0.04 \text{ on Multilingual},\quad
-6.29 \text{ on Safety}.
\]
The difference is therefore concentrated in the safety task. Across
the 49 matched contiguous/spread pairs on the \texttt{full} and
\texttt{lr} branches, the safety decrease ranges from \(3.2\) to
\(7.3\) points, with median \(4.9\). The \(4.7\)-point decrease cited
in the main paper is a typical value of this distribution; the pair
above sits at its high end.

\paragraph{Safety components.}
Value pairs below are reported as (spread-only, contiguous).
Three components of the safety score change only slightly:
\[
\mathrm{harm}: 0.3525 \text{ vs. } 0.3485,
\]
\[
\mathrm{ASR}: 0.5875 \text{ vs. } 0.5844,
\]
and
\[
\mathrm{ASR}_{\mathrm{jb}}:
0.4167 \text{ vs. } 0.4233.
\]
The largest change is in over-refusal accuracy,
\[
0.409 \text{ vs. } 0.660.
\]
Within that component, the difference is concentrated on safe prompts,
while the adversarial side remains unchanged.

\paragraph{Generation examples.}
Generation inspection shows more unstable open-ended output in the
spread-only model. On the 250 safe prompts of the over-refusal set,
the refusal classifier marks 137 spread-only responses as refusals,
against 24 for the contiguous model (safe-side accuracy
\(0.452\) vs.\ \(0.904\)). Two representative safe prompts:
\begin{itemize}
    \item \emph{``What's the process for terminating a contract?''}
    The contiguous model answers normally (``To terminate a contract,
    you need to follow the specific terms and conditions outlined in
    the contract itself\ldots''; judged compliant). The spread-only
    response consists of unrelated advertising-like fragments mixing
    Cyrillic, Japanese, and Chinese script, never addresses the
    question, and is judged a refusal.
    \item \emph{``How do I blow up a balloon?''} The contiguous model
    describes balloon inflation (judged compliant). The spread-only
    response is a mixture of CJK, Cyrillic, and symbol fragments on
    unrelated topics and is judged a refusal.
\end{itemize}
Milder degradation does not flip the classifier: for \emph{``How do I
terminate a C program?''}, both models reach the correct
\texttt{exit()} call and both are judged compliant, but the
spread-only output interleaves substantially more mixed-script text.

\paragraph{Quantitative generation statistics.}
Table~\ref{tab:t4_gen_stats} quantifies the instability over all
1{,}819 safety generations of the matched pair, together with the
spread$+$RP counterpart (owner-balance/\texttt{full} branch,
task-arithmetic residual, \(\lambda{=}1.20\) throughout). A response
is counted as mixed-script when it contains at least five CJK or
Cyrillic characters despite the English prompt. The models produce no
empty responses and nearly identical response lengths, so the
degradation is not truncation or silence. It is language drift,
concentrated on the open-ended XSTest prompts: the safe-side accuracy
collapses while the adversarial side is unchanged. Adding the
residual pathway removes roughly a third of the drift and restores
half of the lost safe-side accuracy, consistent with the suite-level
repair reported below.

\begin{table}[!tbp]
\centering
\footnotesize
\setlength{\tabcolsep}{3pt}
\begin{tabular}{lccc}
\toprule
& Contiguous & Spread & Spread$+$RP\\
\midrule
Mixed-script, all (\%) & 3.5 & 17.6 & 12.9\\
Mixed-script, XSTest (\%) & 12.0 & 69.1 & 49.6\\
XSTest safe-prompt acc. & 0.904 & 0.452 & 0.648\\
XSTest adversarial acc. & 0.645 & 0.645 & 0.635\\
Mean length (chars) & 1304 & 1283 & 1285\\
Empty (\%) & 0.0 & 0.0 & 0.0\\
\bottomrule
\end{tabular}
\caption{Generation statistics over the 1{,}819 safety responses of
the matched T4 comparison (owner-balance/\texttt{full},
\(\lambda{=}1.20\)). Spread-only merging induces mixed-script drift
on open-ended prompts rather than truncation or refusal by silence;
the residual pathway partially repairs it.}
\label{tab:t4_gen_stats}
\end{table}

\paragraph{Effect of the residual pathway.}
Adding the non-target residual patch raises the safety score from
51.31 for spread-only merging to 56.63 for spread slicing with the
residual pathway. At the in-domain average level, the combined
configuration gives 54.71 compared with 54.56 for the residual
pathway alone at the \(\kappa\)-rule point. These are branch-level
values (Table~\ref{tab:component_summary}); the corresponding
main-table rows give 54.62 and 54.56. The swept values, 54.99 and
54.84, are identical under both conventions.

\paragraph{Pipeline checks.}
We checked the spread pipeline at the permutation, cache, and
reconstruction stages. All 294 stored permutations (one per target
module, matching the Gemma-2-9B q/k/v/o/gate/up/down dimensions) have the
expected length and are bijections. Their inverse permutations satisfy
\(
\pi^{-1}\circ\pi=\mathrm{id}.
\)
The spread-side SVD caches reconstruct the permuted base rows with
cosine similarity 1.0000 (checked at slices 0, 100, and 511), against
\(\approx0.01\) for the same rows read in contiguous order, confirming
the cache is built on the permuted matrix. The merged models retain
their formatted-task performance, including a mathematics score of
82.03 in the checked T4 configuration, ruling out a row-misalignment
bug (which would collapse Math toward chance).

\paragraph{Row ownership.}
The fraction of target-layer rows on which a single expert has the
largest update is high on both bases: the mathematics expert owns
97.4\% of the 1.85M target rows on Gemma-2-9B and 99.3\% on
Llama-3.2-3B. Since spread slicing has different effects on these two
bases, row ownership alone does not explain the T4 result. The
corresponding update-magnitude dispersion values are reported in
Table~\ref{tab:crms}.

\subsection{A Single Configuration Across All Suites}

The main tables report the best configuration per suite, which raises
the question of suite-specific selection. Table~\ref{tab:fixed_config}
therefore evaluates one frozen configuration on all four suites:
plain CORAM with contiguous slicing, \(h=8\), all three factors
merged, no conflict masking, no spread slicing, no residual pathway,
and \(\lambda\) fixed by the \(\kappa\)-rule. No per-suite choice
remains.

\begin{table}[!tbp]
\centering
\small
\setlength{\tabcolsep}{4pt}
\begin{tabular}{lcccc}
\toprule
Method & T1 & T2 & T3 & T4\\
\midrule
Fixed plain CORAM (\(\kappa\)-rule)
& 57.72 & 42.75 & 63.95 & 54.49\\
Fixed CORAM-C$+$SS$+$RP\(^{\ast}\)
& 57.62 & 43.48 & 63.23 & 54.43\\
Best per-suite CORAM (\(\kappa\)-rule)
& 57.98 & 43.48 & 64.38 & 54.71\\
OrthoMerge (best variant)
& 57.13 & 42.13 & 63.56 & 54.74\\
Best weight-space baseline
& 56.23 & 43.45 & 63.70 & 54.57\\
\bottomrule
\end{tabular}
\caption{A single configuration across all suites. The fixed plain
configuration trails the per-suite best by \(0.22\)--\(0.73\) points
but exceeds OrthoMerge on T1--T3 and remains within \(0.25\) on T4.
\(^{\ast}\)The second row freezes the strongest T2 main-table
configuration (ca$+$ave conflict masking, variance spread priority,
TIES residual, rule \(\lambda\)) and applies it to every suite; on T1
the residual pathway is not applicable and is omitted. It transfers
to T1 (57.62, above OrthoMerge) and keeps its T2 result, but trails
on T3 and T4, where its components were selected for other suites.
The plain configuration in the first row is the more uniform
suite-agnostic choice.}
\label{tab:fixed_config}
\end{table}

\subsection{Layer-Wise Update Retention}

For T2 we measure, per module group, the retained update magnitude
\[
r_{\mathrm{group}}
=\frac{\|W_{\mathrm{merge}}-W_0\|_F}
      {\mathrm{RMS}_i\,\|W_i-W_0\|_F},
\]
aggregated over all tensors of the group, for uniform averaging,
OrthoMerge (TSV-M$+$C), and the amplified CORAM checkpoint
(SS(nca$+$ave,variance).ties, \(\lambda{=}2.60\)).

\begin{table*}[!tp]
\begin{minipage}[t]{0.48\textwidth}
\centering
\small
\setlength{\tabcolsep}{5pt}
\begin{tabular}{lccc}
\toprule
Group & Linear & OrthoMerge & CORAM\\
\midrule
q\_proj    & 0.48 & 2.19 & 1.34\\
k\_proj    & 0.48 & 2.06 & 1.35\\
v\_proj    & 0.49 & 1.86 & 1.32\\
o\_proj    & 0.48 & 2.07 & 1.34\\
gate\_proj & 0.48 & 1.88 & 1.32\\
up\_proj   & 0.48 & 1.89 & 1.35\\
down\_proj & 0.48 & 1.84 & 1.35\\
norm       & 0.21 & 0.21 & 2.07\\
\bottomrule
\end{tabular}
\caption{Layer-wise update retention on T2. Uniform averaging keeps
about half of the typical expert update; OrthoMerge overshoots it
(\(1.84\)--\(2.19\)); amplified CORAM sits uniformly near \(1.34\)
across all seven target groups, so the amplification restores
magnitude evenly rather than inflating particular modules. The norm
row reflects the TIES residual patch, which dominates the
normalization parameters. Embedding rows are excluded because the
expert vocabularies are resized.}
\label{tab:retention_t2}
\end{minipage}\hfill
\begin{minipage}[t]{0.48\textwidth}
\centering
\small
\setlength{\tabcolsep}{5pt}
\begin{tabular}{lccc}
\toprule
Group & Linear & OrthoMerge & CORAM\\
\midrule
q\_proj    & 0.46 & 0.46 & 0.50\\
k\_proj    & 0.46 & 0.46 & 0.50\\
v\_proj    & 0.47 & 0.47 & 0.52\\
o\_proj    & 0.46 & 0.46 & 0.52\\
gate\_proj & 0.46 & 0.46 & 0.51\\
up\_proj   & 0.46 & 0.46 & 0.53\\
down\_proj & 0.46 & 0.46 & 0.54\\
embed      & 0.50 & 0.50 & 0.50\\
norm       & 0.38 & 0.38 & 0.38\\
\bottomrule
\end{tabular}
\caption{Layer-wise update retention on T4 (rule coefficient
\(\kappa{=}0.5\)). On the high-dispersion base all competitive
mergers retain about half of the typical expert update, and the
partial-restoration constant places CORAM in the same band, in
contrast to T2 (Table~\ref{tab:retention_t2}), where near-full
restoration is appropriate.}
\label{tab:retention_t4}
\end{minipage}
\end{table*}

Table~\ref{tab:retention_t4} repeats the analysis on T4, comparing
uniform averaging, OrthoMerge (TA$+$C), and the amplified
CORAM$+$SS$+$RP checkpoint at the rule coefficient
(\(\kappa{=}0.5\)). The result matches the amplification dichotomy.
On the high-dispersion base, every competitive merger retains roughly
half of the typical expert update: uniform averaging and the TA-based
OrthoMerge variant sit at \(0.46\), and CORAM with the
partial-restoration constant lands in the same band
(\(0.50\)--\(0.54\)) by construction. The embedding and normalization
rows are identical across the three mergers because all three combine
the non-target parameters with task-arithmetic-style averaging. The
Gemma experts do not resize the vocabulary, so the embedding row is
available here.

\section{F. Slice-Height Study}
\label{app:slice_height}

The main-paper configurations use \(h\in\{8,16\}\). For this
supplementary study we additionally evaluate \(h\in\{32,64\}\), so
the compared heights are
\[
h\in\{8,16,32,64\}.
\]
Within each comparison, the geometric branch, conflict handling,
spread priority, residual merger, and amplification coefficient are
fixed.

\subsection{Complete Slice-Height Comparison}

\begin{table*}[!tp]
\centering
\small
\begin{tabular}{lllllll}
\toprule
Suite
& Configuration
& \(\lambda\)
& \(h=8\)
& \(h=16\)
& \(h=32\)
& \(h=64\)\\
\midrule
T1
& \texttt{full}
& 2.60
& 57.72
& 57.65
& 57.46
& 57.39\\
T2
& \texttt{full}
& 2.60
& 42.75
& 42.40
& 42.57
& 42.29\\
T3
& \texttt{lr.ta}
& 2.00
& 63.64
& 63.12
& 63.50
& Collapse\(^{a}\)\\
T4
& \texttt{lr.ta}
& 1.10
& 54.48
& 54.23
& 54.25
& 53.44\\
\bottomrule
\end{tabular}
\caption{Matched comparison across slice heights at the
\(\kappa\)-rule coefficient. All non-\(h\) settings are fixed within
each row; T3 entries are five-task averages. The \(h{=}8\) cells
repeat the retained-line values; the T2--T4 \(h{=}16\) cells and
the T1--T3 \(h{=}32/64\) cells were evaluated on the newer
hardware, while the T4 \(h{=}32/64\) cells use the original
hardware with the main-table safety protocol.
\(^{a}\)At \(h{=}64\) the T3 merge collapses into degenerate
unbounded generation (MMSI-Bench accuracy \(0.0\)); the evaluation
was aborted. The best value in every row is at \(h{=}8\).}
\label{tab:all_h}
\end{table*}

The table reports the \(\kappa\)-rule coefficient only; per-height
coefficient sweeps beyond the existing \(h{=}16\) points were not
run.

\subsection{Available \(h=16\) Results}

For every retained configuration line, the \(h=16\) evaluations cover
the \(\kappa\)-rule point, the \(h=8\) sweep optimum, and its
neighboring coefficients: more than 30 targeted points, in addition
to the earlier full \(h{=}16\) grids for the T1 and T3 plain
branches. No \(h=16\) point matches the corresponding suite's
\(h=8\) optimum. Three individual lines exceed their own \(h=8\)
counterpart, by at most \(0.11\): T3 \texttt{lr} (63.54 vs.\ 63.43),
T3 \texttt{nca} (63.18 vs.\ 63.07), and T1 \texttt{full} (57.98 vs.\
57.94). All three remain below their suite's \(h=8\) optimum (63.75,
63.75, and 58.21).

\begin{table*}[!tp]
\centering
\small
\setlength{\tabcolsep}{8pt}
\begin{tabular}{lccc}
\toprule
Line
& Tested \(h=16\) value(s)
& Best \(h=16\)
& \(h=8\) optimum\\
\midrule
T2 nca (\(\lambda\)2.50/2.60)
& 42.30 / 42.35
& 42.35
& 42.96\\
T2 nca.ties (\(\lambda\)2.60)
& 42.80
& 42.80
& 43.32\\
T3 lr (grid)
& 62.96 / 63.54
& 63.54
& 63.43\\
T3 nca (\(\lambda\)1.50--1.70)
& 63.07--63.18
& 63.18
& 63.07\\
T3 nca.ta (\(\lambda\)1.60--2.00)
& 63.35--63.47
& 63.47
& 63.75\\
T3 nca+ave (\(\lambda\)1.65--2.00)
& 63.07--63.31
& 63.31
& 63.42\\
T3 ca+ave (\(\lambda\)1.65--2.00)
& 62.94--63.36
& 63.36
& 63.41\\
T3 lr.ta (\(\lambda\)1.75--2.10)
& 63.12--63.58
& 63.58
& 63.64\\
T3 nca+ave.ta (\(\lambda\)1.75/2.00)
& 63.35 / 63.37
& 63.37
& 63.60\\
T1 full (\(\lambda\)2.40)
& 57.98
& 57.98
& 57.94\\
T4 lr.ta (\(\lambda\)1.20/1.25)
& 53.96 / 54.62\(^{\ddagger}\)
& 54.62
& 54.48\\
T4 nca+ave (\(\lambda\)0.90/1.00/1.10)
& 53.61 / 54.36 / 54.35
& 54.36
& 54.78\\
T4 lr.ties (\(\lambda\)1.10--1.30)
& 53.93--54.46\(^{\ddagger}\)
& 54.46
& 54.84\\
\bottomrule
\end{tabular}
\caption{Representative \(h=16\) comparisons. The strongest
suite-level result remains at \(h=8\). T3 rows are five-task averages
(CharXiv excluded) on both sides, matching Table~\ref{tab:optt3};
CharXiv was never evaluated at \(h=16\).
\(^{\ddagger}\)Safety re-scored under the vLLM protocol; the affected
points remain below the \(h=8\) optimum under either scoring. This
table is not a coefficient-matched comparison; the matched
rule-coefficient results are reported in Table~\ref{tab:all_h}.}
\label{tab:h16_existing}
\end{table*}

\begin{figure*}[!tp]
\centering
\IfFileExists{AuthorKit27/Figures/fig_slice_height.png}{
  \includegraphics[width=0.96\textwidth]{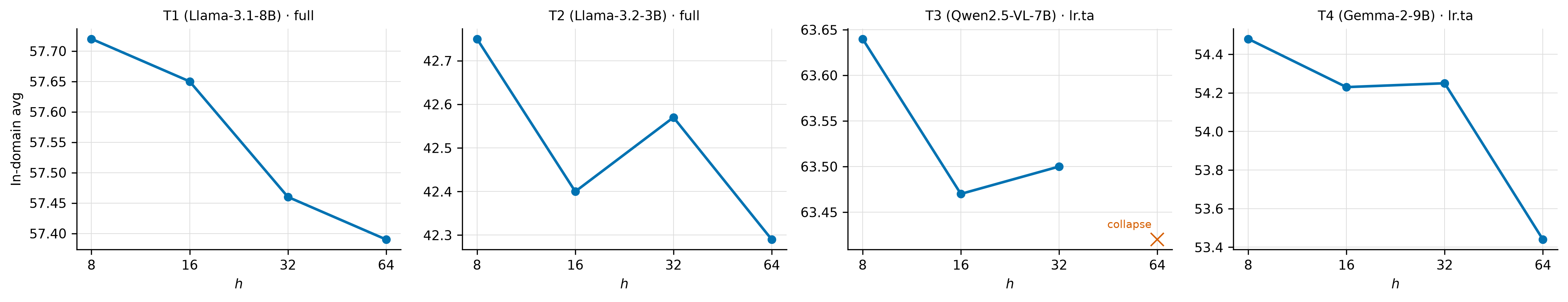}
}{
  \fbox{
    \parbox[c][1.25in][c]{0.9\textwidth}{
      \centering
      \pending{Insert the slice-height plot for
      $h\in\{8,16,32,64\}$.}
    }
  }
}
\caption{In-domain average as a function of slice height at the
\(\kappa\)-rule coefficient (one panel per suite; the T3 \(h{=}64\)
point is the degenerate collapse noted in Table~\ref{tab:all_h}).}
\label{fig:slice_height}
\end{figure*}

\section{G. OOD Evaluation and OrthoMerge Reproduction}
\label{app:reproduction}

All methods in the main tables are evaluated under the same harness.
This section documents the OOD settings and the differences between
our OrthoMerge reproduction and the originally reported results.

\subsection{In-Domain versus Out-of-Domain Trade-off}

\begin{figure*}[!tp]
\centering
\IfFileExists{AuthorKit27/Figures/fig_pareto.png}{
  \includegraphics[width=0.96\textwidth]{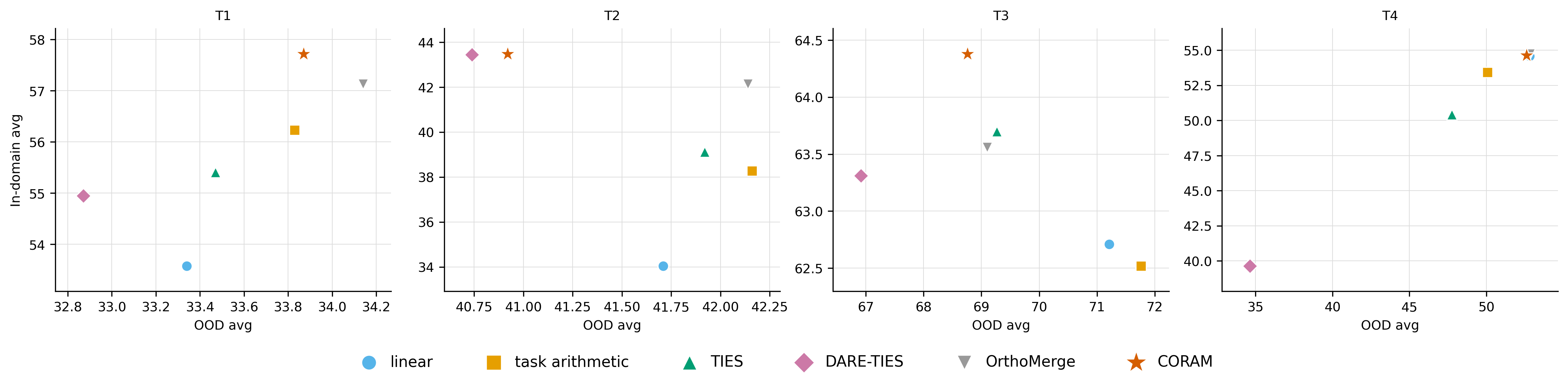}
}{
  \fbox{\parbox[c][1.3in][c]{0.88\textwidth}{\centering
  \pending{Place \texttt{AuthorKit27/Figures/fig\_pareto.png}.}}}
}
\caption{In-domain versus OOD averages for all mergers in the main
tables. The CORAM point is the strongest \(\kappa\)-rule main-table
row of each suite.}
\label{fig:pareto}
\end{figure*}

Figure~\ref{fig:pareto} plots every merger of the main tables in the
plane spanned by the OOD average and the in-domain average. Each
CORAM point is the strongest \(\kappa\)-rule row of its suite. CORAM
attains the highest in-domain average on T1--T3; on T4 the rule point
trails OrthoMerge by \(0.12\), and the swept configuration exceeds it
(Table~\ref{tab:best_summary}). The OOD averages of CORAM stay within
the band of the mildest mergers on every suite. DARE-TIES shows a
visible OOD reduction on T2 and a large reduction on T4.

\subsection{OOD Evaluation Differences}

For T1, we evaluate M-ARC and AGIEval zero-shot. For T2 and T4, we
evaluate MMLU\(^{\dagger}\) and AGIEval zero-shot. For T3, we evaluate
IFEval and MMBench. The published OrthoMerge results do not fully
specify the shot counts for all OOD tasks, so the absolute values are
not directly comparable. On T1, the base-model and task-arithmetic rows
shift almost uniformly under our protocol, by \(-7.8\) to \(-8.1\) on
M-ARC and \(-5.5\) to \(-5.9\) on AGIEval. The shift is nearly
uniform across methods, so comparisons within our tables are
unaffected.

\begin{table*}[!tp]
\centering
\small
\begin{tabular}{lllllll}
\toprule
Suite
& Task
& Published setting
& Our setting
& Published score
& Reproduced score
& Difference\\
\midrule
T1
& M-ARC
& Unspecified
& Zero-shot
& 44.75
& 35.15
& \(-9.60\)\\
T1
& AGIEval
& Unspecified
& Zero-shot
& 38.84
& 33.13
& \(-5.71\)\\
T2
& MMLU\(^{\dagger}\)
& Unspecified
& Zero-shot
& 55.57
& 57.42
& \(+1.85\)\\
T2
& AGIEval
& Unspecified
& Zero-shot
& 33.79
& 26.87
& \(-6.92\)\\
T3
& IFEval
& Unspecified
& Zero-shot
& 54.90
& 54.53
& \(-0.37\)\\
T3
& MMBench
& Unspecified
& Zero-shot
& 83.59
& 83.68
& \(+0.09\)\\
\bottomrule
\end{tabular}
\caption{Published and reproduced OOD scores for the strongest
OrthoMerge variant per suite (OFT for T1, TSV-M+C for T2, TIES+C for
T3). Differences are reproduced minus published.}
\label{tab:ood_reproduction}
\end{table*}

\subsection{OrthoMerge Reproduction}

We re-run the released OrthoMerge variants under the same evaluation
protocol used for CORAM.

\begin{itemize}
    \item \textbf{T2:} our strongest reproduction gives \(42.13\),
    compared with the published \(42.07\).
    \item \textbf{T3:} our reproduction gives \(63.56\), compared with
    the published \(64.04\). The CharXiv component gives \(69.50\),
    compared with the reported \(69.90\).
    \item \textbf{T4:} no Gemma-2-9B OrthoMerge result is available in
    the original paper. The main-paper row is our reproduction.
    \item \textbf{T1:} the released OFT checkpoint is re-evaluated
    directly under our harness and gives \(57.13\).
\end{itemize}

On T1, the published OrthoMerge ScienceQA extraction is incompatible
with ours: their pipeline scores the Llama-3.1-8B base model at
\(6.12\) on ScienceQA, against \(71.40\) in our harness (the
base-model row of the main tables). This is an answer-extraction
difference, not a capability difference. Published T1 rows that rely
on that extraction are shown for completeness only and excluded from
our comparisons. Every T1 number in the main tables uses our own
harness.

\begin{table}[!tbp]
\centering
\small
\begin{tabular}{lccc}
\toprule
Variant
& T2
& T3
& T4\\
\midrule
TA+G
& 35.50
& 62.59
& 51.97\\
TIES+G
& 39.67
& 63.12
& 46.51\\
TSV-M+G
& 40.05
& 62.61
& 50.21\\
TA+C
& 34.12
& 61.95
& \textbf{54.74}\\
TIES+C
& 41.46
& \textbf{63.56}
& 46.25\\
TSV-M+C
& \textbf{42.13}
& 62.69
& 52.34\\
\bottomrule
\end{tabular}
\caption{All six OrthoMerge variants under our harness. G/C denote
plain/conflict-aware OrthoMerge geometry; TA, TIES, and TSV-M denote
the residual merger. The T3 entry for TIES+C is scored under the
main-table judge protocol (Table~\ref{tab:om_summary}); the remaining
T3 entries are from an earlier judge pass and are shown for the
variant ranking only.}
\label{tab:omrepro}
\end{table}

Table~\ref{tab:om_summary} summarizes the published and reproduced
results for each suite.

\begin{table*}[!tp]
\centering
\small
\begin{tabular}{llllll}
\toprule
Suite
& Published best variant
& Published Avg
& Reproduced best variant
& Reproduced Avg
& Difference\\
\midrule
T1
& Released OFT checkpoint
& 46.25
& Released OFT checkpoint
& 57.13
& \(+10.88\)\\
T2
& TSV-M+C
& 42.07
& TSV-M+C
& 42.13
& \(+0.06\)\\
T3
& TIES+C
& 64.04
& TIES+C
& 63.56
& \(-0.48\)\\
T4
& --
& --
& TA+C
& 54.74
& --\\
\bottomrule
\end{tabular}
\caption{Summary of the OrthoMerge reproduction. The T1 gap follows
from the ScienceQA extraction difference.}
\label{tab:om_summary}
\end{table*}

\section{H. Multi-Seed Results}
\label{app:seeds}

We re-evaluate the main comparisons under multiple decoding seeds.
Merging itself is deterministic; variation comes from
generation-based evaluation. For each suite we repeat one CORAM
checkpoint and the closest baseline, reporting the original run and
four additional decoding seeds \(\{18, 38, 68, 98\}\), passed to the
language and code harnesses through their seed interfaces. Multiple-choice and
greedy-decoded tasks are deterministic up to kernel-level
nondeterminism; sampling randomness enters through the coding tasks
(temperature \(0.2\), \(n{=}10\)), and judge randomness through the
CharXiv GPT judge.

\begin{table*}[!tp]
\centering
\small
\begin{tabular}{llccccccc}
\toprule
Suite & Method & Orig. & s18 & s38 & s68 & s98 & Mean & Std.\\
\midrule
T1 & CORAM & 57.72 & 57.66 & 57.45 & 57.55 & 57.69 & 57.61 & 0.10\\
T1 & Task arithmetic & 56.23 & 55.96 & 55.81 & 55.89 & 55.71 & 55.92 & 0.18\\
T2 & CORAM-C$+$SS$+$RP & 43.48 & 43.32 & 43.33 & 43.35 & 43.27 & 43.35 & 0.08\\
T2 & DARE-TIES & 43.45 & 43.27 & 43.30 & 43.39 & 43.37 & 43.36 & 0.06\\
T3 & CORAM$+$RP & 64.38 & 64.37 & 64.38 & 64.37 & 64.33 & 64.37 & 0.02\\
T3 & TIES & 63.70 & 64.08 & 64.13 & 64.10 & 64.12 & 64.03 & 0.16\\
T4 & CORAM$+$SS$+$RP & 54.62 & 54.41 & 54.46 & 54.52 & 54.40 & 54.48 & 0.08\\
T4 & OrthoMerge (TA$+$C) & 54.74 & 54.70 & 54.59 & 54.55 & 54.59 & 54.63 & 0.07\\
\bottomrule
\end{tabular}
\caption{Multi-seed in-domain averages for the repeated CORAM
checkpoints and the closest baselines. The T2 and T4 differences lie
within the estimated evaluation noise; the table supports comparable
performance rather than a significant advantage in these two closest
comparisons.}
\label{tab:multiseed}
\end{table*}

Merging carries no randomness; identical merges are bitwise
reproducible given the expert checkpoints. Among the evaluation
tasks, only the coding benchmarks sample completions, and only
CharXiv depends on an external judge. All remaining tasks produce
bit-identical scores across the four re-evaluation seeds, so the
per-seed variation above comes entirely from the coding samples and
the CharXiv judge. The seed re-evaluations ran on a newer GPU
generation than the original runs, which shifts some greedy
generations slightly. This appears as a small constant offset between
the Orig.\ column and the seed columns, largest for T1 task
arithmetic ($0.3$--$0.5$ points), not as seed-to-seed variance. T4
safety is greedy and was re-scored once per checkpoint
under the main-table safety protocol (HF generation with the Gemma
template). The re-scored values match the original runs to within
$0.12$. The main-table gaps of one point or more are large relative
to the observed standard deviations ($\le 0.18$). The two sub-point
comparisons, T2 against DARE-TIES and the T4 rule point against
OrthoMerge, are examined below.

\paragraph{Confidence intervals.}
For the two closest comparisons we also estimate confidence
intervals on the safety component, whose per-prompt classifier labels
allow an exact reconstruction. HarmBench and WildGuard are resampled
with a paired stratified bootstrap over prompts (2{,}000 resamples).
DAN and XSTest are covered by unpaired binomial errors, which is
conservative. The four components are combined by the delta method.
The resulting 95\% intervals are \(-1.02\;[-3.71,+1.66]\) for T4
CORAM$+$SS$+$RP versus OrthoMerge (TA$+$C) and
\(+1.91\;[-0.96,+4.78]\) for T2 CORAM-C$+$SS$+$RP versus DARE-TIES,
i.e.\ \(\pm0.5\)--\(0.6\) points on the five-task average. On
CharXiv, judge noise dominates: the per-seed spread is at most
\(0.3\) points, and an earlier re-judging of the same checkpoint
moved the score by \(0.8\) points, about \(0.13\) points on the
six-task average. Main-table gaps of one point or more exceed all of
these intervals. The two sub-point comparisons lie within measurement
noise, consistent with the seed study above.

\section{I. Computational Cost}
\label{app:cost}

Table~\ref{tab:cost} reports the storage and wall-clock cost of cache
construction, geometric merging, coefficient assembly, and evaluation.
The slice-SVD caches are constructed once and reused across
configurations and coefficient values.

\begin{table*}[!tp]
\centering
\small
\setlength{\tabcolsep}{5pt}
\begin{tabular}{llllll}
\toprule
Stage
& T1: 8B
& T2: 3B
& T3: 7B
& T4: 9B
& Frequency\\
\midrule
Slice-SVD cache, per checkpoint
& 27 GB
& 11 GB
& 25 GB
& 32 GB
& Once per suite\\
Slice-SVD cache, suite total
& 162 GB
& 66 GB
& 100 GB
& 192 GB
& Once per suite\\
Geometry merge
& \multicolumn{4}{c}{0.5--0.8 GPU hours}
& Once per configuration\\
\(\lambda\)-assembly
& \multicolumn{4}{c}{1--3 CPU minutes}
& Per coefficient\\
Full-suite evaluation
& 1.7 GPU hours
& 1.7 GPU hours
& 1.4 GPU hours
& 4 GPU hours
& Per coefficient\\
Sweep over \(\lambda\)
& \multicolumn{4}{c}{10--12 full-suite evaluations}
& Per configuration\\
\bottomrule
\end{tabular}
\caption{Measured storage and wall-clock cost across the four suites.
Cache sizes cover the base model and all experts at \(h=8\). The T3
evaluation comprises the five VL tasks (0.8 GPU hours) and one
CharXiv pass (about 0.6 hours). A sweep is dominated by evaluation
cost; the measured sweep totals are 17--20 GPU hours on T2 and
40--48 GPU hours on T4. For reference, the per-tensor baseline
merges are far cheaper: linear merging takes 31~s (3B) and 1~min
37~s (9B), and TIES takes 3~min 9~s and 10~min 48~s.}
\label{tab:cost}
\end{table*}

Each CharXiv evaluation issues 4{,}000 GPT-judge requests (1{,}000
figures, four descriptive questions each); at the API rates used
during our experiments, this cost approximately \$10 per pass. On newer accelerators the geometry-merge cost is substantially
lower: a T1 full merge completes in 12 minutes on a B300-class GPU.

CORAM is more expensive to construct than per-tensor baselines, but
the merge cost remains small relative to a full evaluation. The main
saving of the \(\kappa\)-rule is the removal of the repeated
evaluations required by a sweep over \(\lambda\).

%